\documentclass[10pt,twocolumn,letterpaper]{article}

\usepackage[pagenumbers]{cvpr} % To force page numbers, e.g. for an arXiv version

\usepackage{booktabs}
\usepackage{graphicx}
\usepackage{pifont}
\usepackage{multirow}
\usepackage{amsmath}
\usepackage{amssymb}
\usepackage{algorithm}
\usepackage{algpseudocode}
\usepackage{makecell}
\usepackage{subcaption}
\usepackage[table]{xcolor}
\usepackage[most,skins,theorems]{tcolorbox}
\definecolor{deepiceblue}{RGB}{70,150,230}
\definecolor{deeppinkmark}{RGB}{220,80,150}
\definecolor{light_blue}{HTML}{ECF4FF}
\definecolor{light_pink}{HTML}{FDEEF4}

\newcommand{\cmark}{\textcolor{deepiceblue}{\ding{51}}}
\newcommand{\xmark}{\textcolor{deeppinkmark}{\ding{55}}}

\newtcolorbox{myboxi}[1][]{
  breakable,
  colback=purple!5,
  colbacktitle=purple!5,
  coltitle=black,
  fonttitle=\bfseries,
  bottomrule=0pt,
  toprule=0pt,
  leftrule=2pt,
  rightrule=2pt,
  titlerule=0pt,
  arc=0pt,
  outer arc=0pt,
  colframe=cvprblue!60,
  before skip=5pt,
  after skip=0pt,
}
\definecolor{cvprblue}{rgb}{0.21,0.49,0.74}
\usepackage[pagebackref,breaklinks,colorlinks,allcolors=cvprblue]{hyperref}
\usepackage[capitalize,nameinlink,noabbrev]{cleveref}

\def\paperID{10092} % *** Enter the Paper ID here
\def\confName{CVPR}
\def\confYear{2026}

\title{Video-R4: Reinforcing Text-Rich Video Reasoning with Visual Rumination}

\author{\textbf{Yolo Yunlong Tang}$^1$, \textbf{Daiki Shimada}$^2$, \textbf{Hang Hua}$^3$, \textbf{Chao Huang}$^1$, \textbf{Jing Bi}$^1$,\\\textbf{Rogerio Feris}$^3$, \textbf{Chenliang Xu}$^1$\\
$^1$University of Rochester, $^2$Sony Group Corporation, $^3$MIT-IBM Watson AI Lab\\
\\
\url{https://yunlong10.github.io/Video-R4/}
}

\begin{document}
\maketitle

\begin{abstract}
Understanding text-rich videos requires reading small, transient textual cues that often demand repeated inspection. Yet most video QA models rely on single-pass perception over fixed frames, leading to hallucinations and failures on fine-grained evidence. Inspired by how humans pause, zoom, and re-read critical regions, we introduce Video-R4 (\textbf{R}einforcing Text-\textbf{R}ich Video \textbf{R}easoning with Visual \textbf{R}umination), a video reasoning agent that performs visual rumination: iteratively selecting frames, zooming into informative regions, re-encoding retrieved pixels, and updating its reasoning state. We construct two datasets with executable rumination trajectories: Video-R4-CoT-17k for supervised practice and Video-R4-RL-30k for reinforcement learning. We propose a multi-stage rumination learning framework that progressively finetunes a 7B LMM to learn atomic and mixing visual operations via SFT and GRPO-based RL. Video-R4-7B achieves state-of-the-art results on M4-ViteVQA and further generalizes to multi-page document QA, slides QA, and generic video QA, demonstrating that iterative rumination is an effective paradigm for pixel-grounded multimodal reasoning.
\end{abstract}    
\begin{figure}
    \centering
    \includegraphics[width=\linewidth]{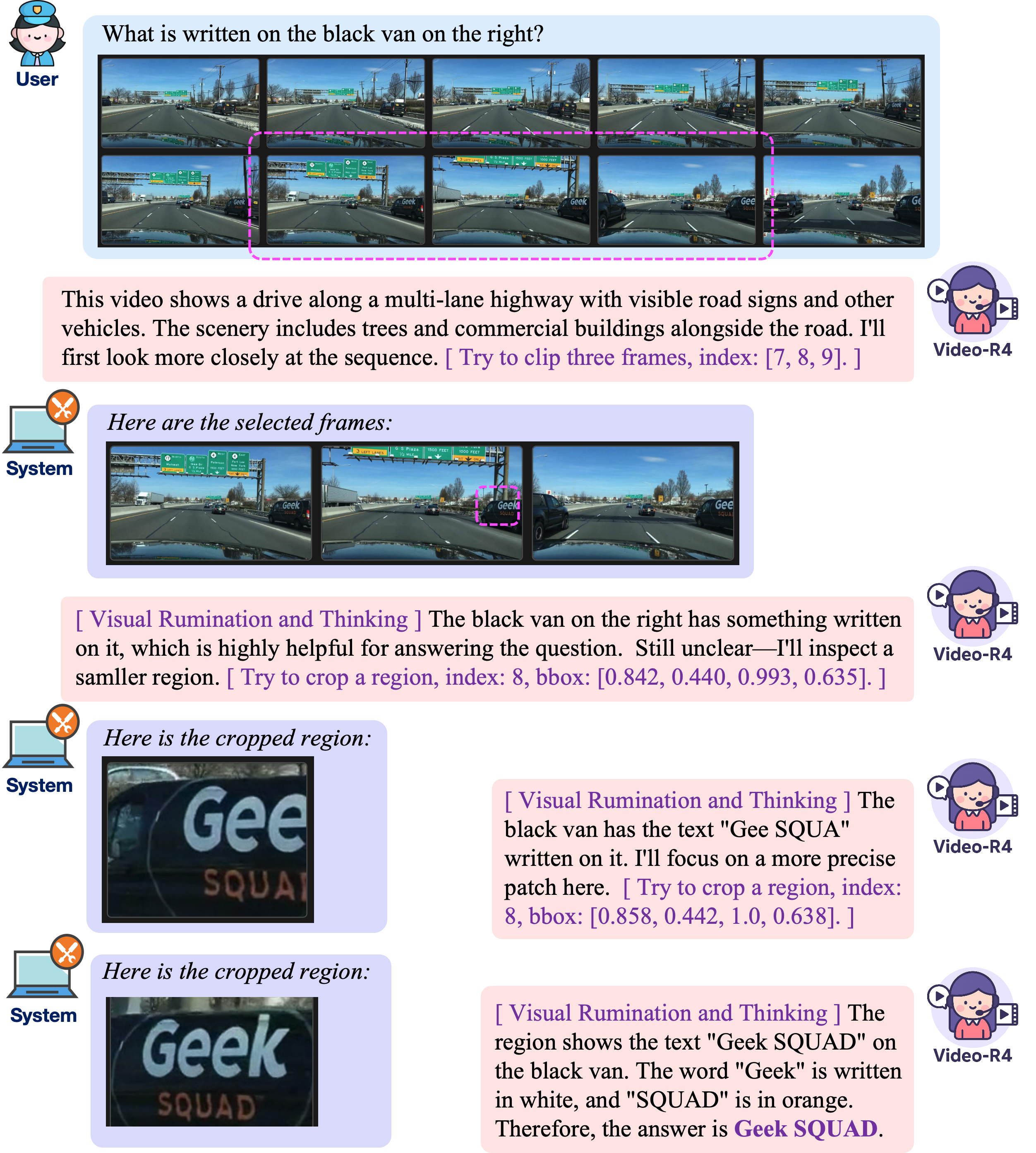}
    \vspace{-0.3cm}
    \caption{Video-R4 performs iterative visual rumination by selecting frames, zooming into regions, and re-encoding pixels, forming a closed-loop \textbf{read–retrieve–refocus–reinforce} cycle for grounded video reasoning.}
    \label{fig:teaser}
    \vspace{-1em}
\end{figure}
\begin{figure*}[!ht]
    \centering
    \includegraphics[width=\linewidth]{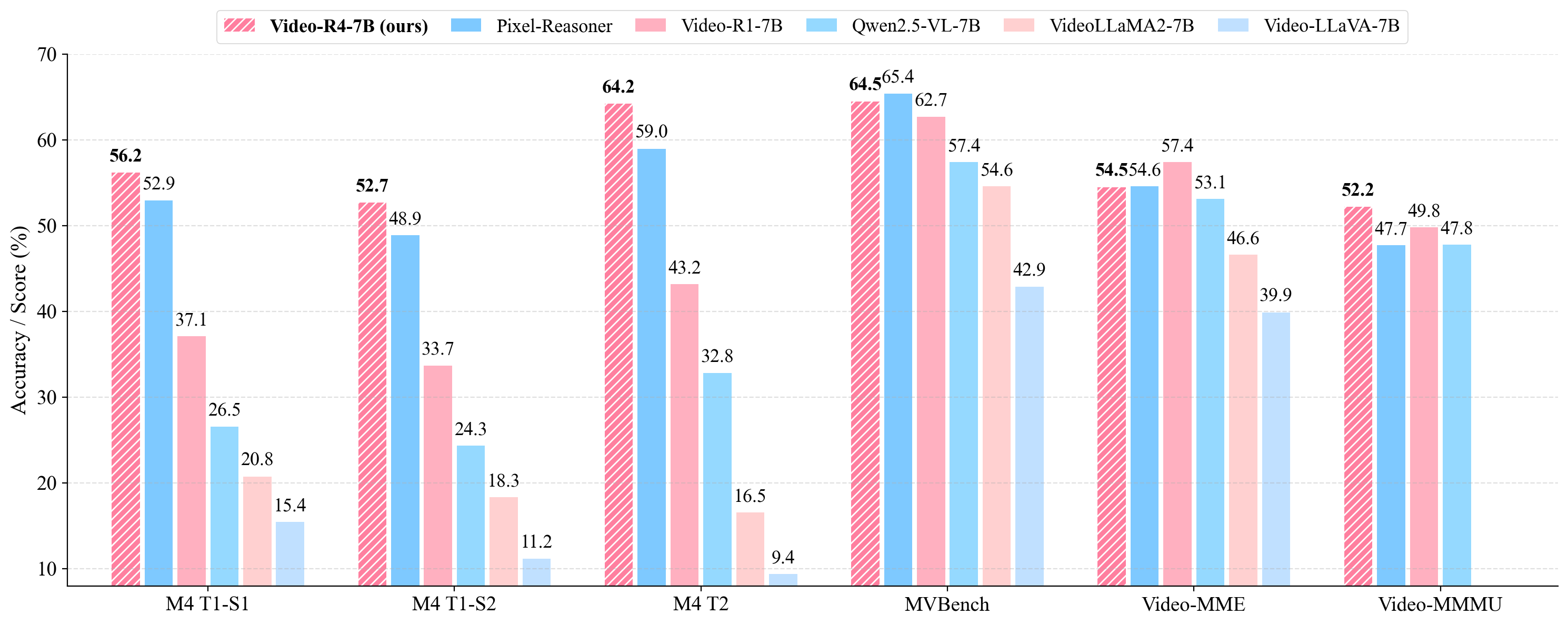}
    \vspace{-0.7cm}
    \caption{Our Video-R4-7B model achieves state-of-the-art performance on the text-rich video understanding dataset M4-ViteVQA, and is also compatible with the LMMs with the same size on the general video QA benchmarks.}
    % \vspace{-0.5cm}
    % \caption{Overall performance comparison on the M4-ViteVQA benchmark. Our \textbf{Video-R4-7B} achieves \textbf{state-of-the-art accuracy and ANLS} across all evaluation splits, outperforming previous video LMMs, visual-grounded systems, and RL-enhanced models of comparable size. The closed-loop rumination mechanism, iterative frame selection, spatial zooming, and pixel re-encoding, substantially improves grounding in text-rich videos and yields consistent gains over the Qwen2.5-VL backbone.}
    \label{fig:bar}
\end{figure*}
\vspace{-0.2cm}
\section{Introduction}

Understanding text-rich videos requires precise reading of small, transient textual cues that often appear only in specific frames or localized regions. Recent video question answering and text-centric video benchmarks have highlighted these challenges in news videos, driving scenes, egocentric recordings, and UI or slide walkthroughs~\cite{zhao2022vitevqa,jahagirdar2023watching,tom2023reading,liu2025gesturelsmlatentshortcutbased,contexualliu2025,zhou2025egotextvqa,zhang2025gatherandtrace,hua2024mmcomposition}, while broader surveys on video understanding with large multimodal models (LMMs) underline the difficulty of scaling such capabilities to long, complex videos~\cite{tang2025video}. Despite advances in video-focused LMMs and long-video benchmarks~\cite{wang2023allinoneb,videochatonline,li2024mvbench,fu2025videomme,hu2025videommmu,liu2024oryx,shu2024video,liu2025video,yuan2025memory}, most systems operate under a \emph{single-pass perception paradigm}, processing a fixed set of frames and relying heavily on text-only chain-of-thought to fill in missing details.
This design leads to brittle behavior in text-rich scenarios. Once a set of frames has been selected and encoded, the model typically cannot revisit frames, re-examine regions, or refine beliefs when initial perceptions are incomplete. Text-only chain-of-thought prompting can improve reasoning~\cite{wei2022chain,zhou2022least,zhang2023multimodal,chen2025towards,li2025system,tang2025vidcomposition,tang2025mmperspective,hua2025finecaption}, but when predictions are not grounded in pixels, it can also amplify hallucinations about content that was never observed. Meanwhile, coordinate-grounded approaches in TextVQA, TextVideoQA, and document VQA predict frame indices, bounding boxes, or layout regions as intermediate evidence~\cite{hu2020iterative,zhou2023exploring,zhou2024graph,jahagirdar2023watching,tanaka2021visualmrc,tito2023hierarchical,tanaka2023slidevqa,huang2022layoutlmv3}, yet these coordinates are usually treated as static endpoints rather than actionable instructions: the referenced pixels are rarely brought back into the model’s context to be re-read and compared.

In contrast, human viewers naturally adopt an iterative ``pause-zoom-check'' strategy when watching text-heavy videos such as screen recordings, lecture slides, or UI demos. We pause at a relevant moment, zoom into a region, reread the text, compare across frames, and revise our understanding as new evidence emerges. This observation forms the core inspiration for our work: if an LMM were equipped with the ability to act on the video, select frames, zoom into regions, fetch higher-resolution pixels, and incorporate them back into its context, it could escape the limitations of one-shot perception and move toward pixel-grounded, multi-step reasoning.

Motivated by this, we propose \textbf{Video-R4} (\textbf{R}einforcing Text-\textbf{R}ich Video \textbf{R}easoning with Visual \textbf{R}umination), a video reasoning LMM that performs \emph{visual rumination}. As shown in \cref{fig:teaser}, the model executes cycles of selecting informative frames, zooming into fine-grained regions, re-encoding the retrieved pixels, and updating its internal reasoning state. This closed-loop---\emph{read, retrieve, refocus, reinforce}---turns temporal selection and spatial zoom into explicit decision steps, allowing the model to accumulate and verify evidence across multiple iterations rather than relying on a single perception of the video. Our design is complementary to recent RL-based reasoning efforts in language/multimodal models~\cite{guo2025deepseek,jaech2024openai,li2024process,chen2025towards,li2025system,r1vl,videor1,huang2025vision,visualrft,segzero,fan2025sophiavl,song2024texttoon,song2025streamme}, but specifically targets text-rich video reasoning with an explicit control interface for visual operations.

\begin{figure*}
    \centering
    \includegraphics[width=1\linewidth]{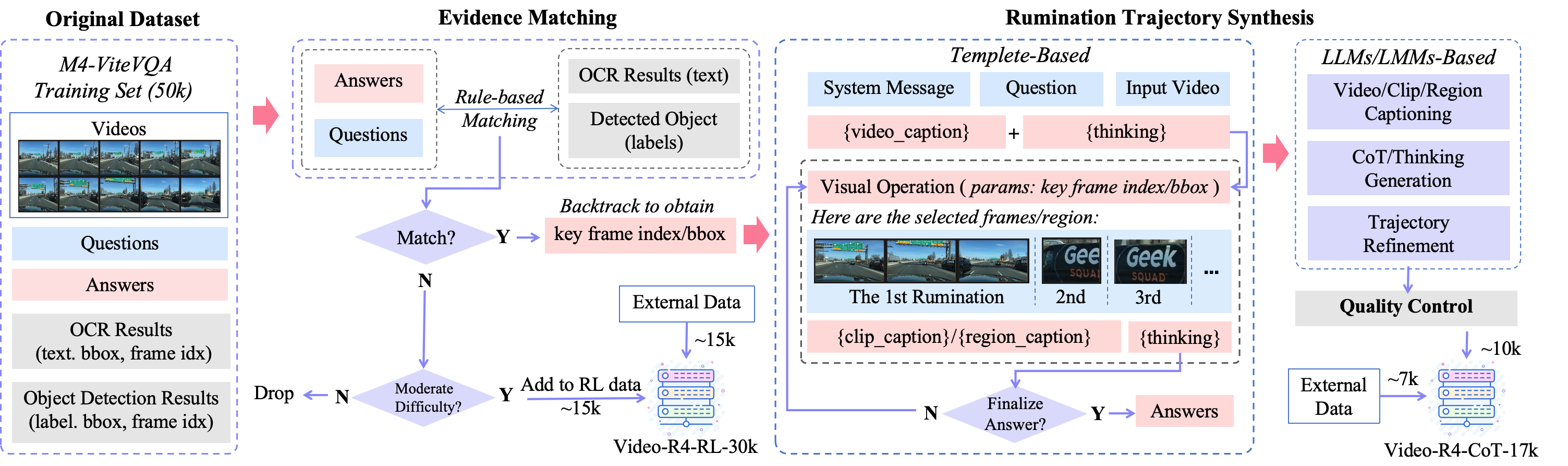}
    \caption{Data curation pipeline for creating the Video-R4-CoT-17k for supervised deliberate rumination practice fine-tuning (DRP-SFT) and compositional rumination practice fine-tuning (CRP-SFT), as well as the Video-R4-RL-30k dataset for reinforcement learning. The \colorbox{light_blue}{light blue parts} are intended to be used as the model’s inputs, while the \colorbox{light_pink}{pink parts} are expected to be produced by the model as outputs.}
    \label{fig:data_pipeline}
    \vspace{-0.3cm}
\end{figure*}

Training such behavior is nontrivial: multi-step rumination requires not only learning \emph{how} to use visual operations, but also \emph{when} and \emph{why} to apply them. To this end, we curate executable trajectories from the M4-ViteVQA dataset~\cite{zhao2022vitevqa} and design a \emph{multi-stage rumination learning framework} that progressively teaches atomic and compositional operations via GRPO-style reinforcement learning built on PPO~\cite{schulman2017proximal}. Our reward design draws on ideas from diversity- and representativeness-based video summarization~\cite{zhou2018deep} and recent curiosity-driven and vision-centric reinforcement learning for pixel-space reasoning~\cite{su2025pixelreasoner,visualrft}. This staged curriculum emerges as a strong inductive bias, yielding faster convergence and significantly higher final performance than collapsed or single-stage methods. Empirically, Video-R4 sets a new state of the art on M4-ViteVQA~\cite{zhao2022vitevqa} and generalizes well beyond its training domain. Despite being trained exclusively on text-rich videos, the model transfers effectively to multi-page document QA and slides QA benchmarks~\cite{tanaka2021visualmrc,tito2023hierarchical,tanaka2023slidevqa,huang2022layoutlmv3}, as well as to general video QA benchmarks such as MVBench, Video-MME, and Video-MMMU~\cite{li2024mvbench,fu2025videomme,hu2025videommmu}. These results suggest that iterative visual rumination forms a broadly useful paradigm for multimodal reasoning over both videos and long documents. In summary, our contributions are:
\begin{itemize}
\item We construct two curated datasets for executable text-rich video reasoning: \textbf{Video-R4-CoT-17k} for supervised rumination pratice and \textbf{Video-R4-RL-30k} for reinforcement learning, enabling study of temporal selection, spatial zooming, and multi-step evidence acquisition.
\vspace{4pt}

\item We introduce \textbf{Video-R4}, a video reasoning LMM that performs iterative visual rumination by selecting frames, zooming into regions, re-encoding pixels, and updating its reasoning state, and continually updating its internal state to support pixel-grounded reasoning.
\vspace{4pt}

\item We develop a \textbf{multi-stage rumination learning framework} that incrementally teaches atomic operations, compositional sequences, and operation control via GRPO-based reinforcement learning. This staged curriculum yields faster convergence and substantially stronger final performance than single-stage or collapsed alternatives.
\vspace{4pt}

\item We achieve \textbf{state-of-the-art performance on M4-ViteVQA} and demonstrate robust transfer to multi-page document QA, slides QA, and general video QA, highlighting the broad applicability of iterative visual rumination beyond the training domain.
\end{itemize}
\section{Method: Video-R4}
\subsection{Data Curation}

\paragraph{Data Source.}
As shown in \cref{fig:data_pipeline}, we start from the training split of M4-ViteVQA, a text-rich VideoQA benchmark with more than fifty thousand question-answering pairs and diverse real-world scenes \cite{zhao2022vitevqa}. Each sample provides a video, one question, and its answer, together with pre-extracted OCR tokens and object detection results. These annotations retain text content, frame indices, object labels, and bounding boxes, which form the evidence pool for rumination trajectory synthesis.

\paragraph{Evidence Matching.}
We aim to recover the evidence needed to answer each question and thereby prepare explicit chains of thought. Starting from the gold answer, we apply rule-based string matching between answers and OCR tokens, as well as between linguistic mentions of entities and object labels. We adopt fuzzy matching with edit distance to handle recognition noise and minor wording variation \cite{levenshtein1966binary}. For matched samples, we record supporting frame indices and bounding boxes and mark whether the matched text or object is truly helpful for solving the question. For unmatched samples, we estimate question difficulty and keep moderately difficult cases as candidates for Video-R4-RL-30k and drop the rest of the samples, which can make the GRPO-based RL more stable. The full matching rules details are given in the appendix.

\paragraph{Rumination Trajectory Synthesis.}
Given matched evidence, we synthesize rumination trajectories using a lightweight chain-of-thought template. The template interleaves internal thinking steps and visual operations applied to the video. We define two atomic visual operations inspired by recent work on video reasoning \cite{videor1,chen2025towards}. \emph{Clipping} selects several key frames by their indices and prompts the model to describe each frame. \emph{Cropping} focuses on one frame and extracts a salient region with a bounding box, followed by a region-level caption. All frames and regions are restricted to those produced by the matching stage, ensuring that every step remains grounded in observed evidence. We then fill the template using a strong video-capable multimodal model and further check temporal consistency and answer correctness \cite{bai2025qwen25,chen2024far}. Valid trajectories become supervision for the chain of thought dataset Video-R4-CoT-17k.

\paragraph{Quality Control.}
We develop an annotation interface that displays each trajectory alongside visualized key frames, cropped regions, and the corresponding question-answer pair. Annotators quickly scan the rumination steps, verify that every visual operation points to the right evidence, and confirm that the final answer follows from the collected clues. Samples with hallucinated or missing evidence are edited or removed. Beyond M4-ViteVQA, we gather additional text-centric VideoQA instances from public datasets and convert them into the same format \cite{tom2023reading,jahagirdar2023watching,zhou2025egotextvqa,su2025pixelreasoner}. After automatic and manual filtering, we obtain about 17k trajectories for Video-R4-CoT-17k and about 30k reinforcement learning samples for Video-R4-RL-30k.

\subsection{Preliminary of GRPO}
We adopt Group Relative Policy Optimization (GRPO)~\cite{guo2025deepseek} as the core policy optimization algorithm. Firstly, the policy $\pi_\theta$ samples a group of $G$ distinct candidate responses (or trajectories) $\{o_1, \dots, o_G\}$ for a given input query $q$. After calculating with predefined reward functions, we obtain their corresponding rewards $\{R_1, \dots, R_G\}$. We compute the group-wise mean and standard deviation, and define the relative quality of the $i$-th response as:
\begin{equation}
A_i = \frac{R_i - \mathrm{mean}(\{R_j\}^G_{j=1})}{\mathrm{std}(\{R_j\}^G_{j=1})}
\end{equation}
The policy is optimized to increase the probability of actions with higher group-relative advantage and decrease those with lower advantage. Following PPO~\cite{schulman2017proximal}, we apply a clipped objective to stabilize updates:
\begin{equation}
\begin{aligned}
&\mathcal{J}_{\mathrm{GRPO}}(\theta)
= 
\mathbb{E}_{q, \{o_i\}} \Bigg[
\frac{1}{G} \sum_{i=1}^{G}
\Bigg(
\min\Bigg(
\frac{\pi_\theta(o_i \mid q)}{\pi_{\theta_{\text{old}}}(o_i \mid q)} A_i, \\[2pt]
&\mathrm{clip}\Bigg(
\frac{\pi_\theta(o_i \mid q)}{\pi_{\theta_{\text{old}}}(o_i \mid q)},
1 - \epsilon,\,
1 + \epsilon
\Bigg) A_i
\Bigg)
- \gamma\, \mathbb{D}_{\mathrm{KL}}(\pi_\theta \,\|\, \pi_{\mathrm{ref}})
\Bigg],
\end{aligned}
\end{equation}
where $\mathbb{D}_\mathrm{KL}$ is KL-divergence term to prevent the optimized policy $\pi_\theta$ from far from the original LMM $\pi_\text{ref}$, and $\gamma$ is a regularization coefficient.
This group-relative formulation reduces variance compared to individual-sample policy gradient updates, improves optimization robustness, and encourages relative action ranking rather than relying solely on absolute reward magnitudes.

\subsection{Reward Design}
\label{sec:rewards}
Our reward function combines four components: the original reward $R$ (\eg, answer correctness and format), Diversity Reward $R_\text{div}$, Representativeness Reward $R_\text{rep}$, and Curiosity Reward $R_\text{cur}$. The overall reward is:

\begin{equation}
    R' = R+\lambda_{\text{div}} R_{\text{div}} + \lambda_{\text{rep}} R_{\text{rep}} + \lambda_{\text{cur}} R_{\text{cur}},
\end{equation}
where the choice of the coefficients, $\lambda_\text{div}$, $\lambda_\text{rep}$, and $\lambda_\text{cur}$, can be found in the appendix (\Cref{app:sec:training_details}).

\paragraph{Diversity Reward.}
Following prior unsupervised summarization objectives~\cite{zhou2018deep}, we encourage selected regions to be mutually dissimilar in feature space. Let $V$ denote the set of features of the input frames and $\hat{V}=\hat{V}^{f}\cup\hat{V}^{r}$ denote the set of features of the selected frames ($\hat{V}^{f}$) and regions ($\hat{V}^{r}$), where $\hat{V}^{f}\subseteq V$. We define representativeness reward to encourage the policy to avoid redundant region selections:
\begin{equation}
R_{\mathrm{div}}(\hat{V}^{r}) = 
\frac{1}{|{\hat{V}^{r}}| (|{\hat{V}^{r}}| - 1)} 
\sum_{i=1}^{|\hat{V}^{r}|} \sum_{j \ne i}^{|\hat{V}^{r}-1|} 
d(v_i, v_{j}),
\end{equation}
where $d(\cdot,\cdot)$ denotes cosine similarity: 
\begin{equation}
d(v_i, v_{j}) = 1 - \frac{v_i^{\top} v_{j}}{\|v_i\|_2 \|v_{j}\|_2}.
\end{equation}
This objective computes the average pairwise distance between all selected region features in $\hat{V}^r$. The normalization by $|\hat{V}^r| (|\hat{V}^r|-1)$ makes the scale of $R_{\mathrm{div}}$ comparable across different numbers of selected regions, so the policy is not trivially rewarded for increasing $|\hat{V}^r|$. Using cosine-based distance further makes the reward depend on the orientation of features rather than their magnitude, which aligns it with semantic differences captured by the encoder. As a result, $R_{\mathrm{div}}$ biases the training dynamics toward solutions where the selected crops are spread out in the feature space instead of collapsing to a narrow cluster.

\begin{figure*}
    \centering
    \includegraphics[width=1\linewidth]{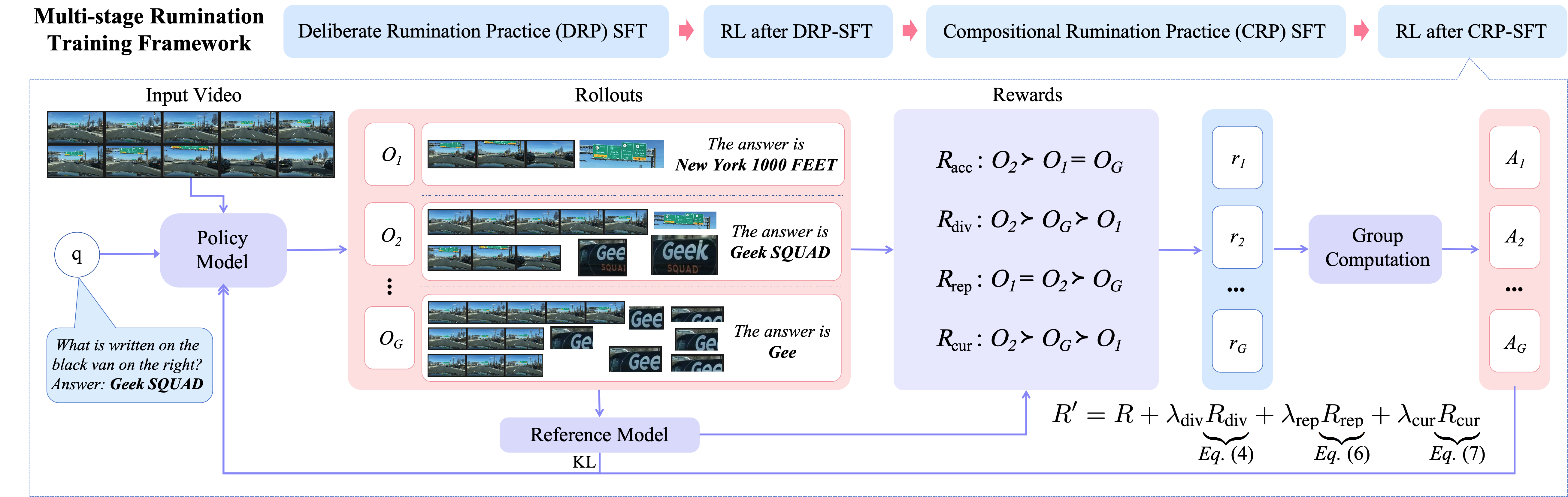}
    \caption{Overview of multi-stage rumination training framework.}
    \label{fig:training_framework}
    
\end{figure*}

\paragraph{Representativeness Reward.}
To ensure the selected frames $\hat{V}^{f}$ remain informative, we encourage them to represent the global video frame set $V$ with the representativeness reward:
\begin{equation}
R_{\mathrm{rep}}(V,\hat{V}^f) = 
\exp{\!\left(
-\frac{1}{|V|} 
\sum_{i=1}^{|V|} 
\min_{v_j \in \hat{V}^{f[-1]}} 
\| v_i - v_{j} \|_2
\right)},
\end{equation}
where $\|\cdot\|_2$ denotes the Euclidean distance in the feature space and $\hat{V}^{f[-1]}$ is the set of frame features selected in the last clipping operation. To reduce computational redundancy, we only use $\hat{V}^{f[-1]}$ to compute $R_{\mathrm{rep}}$.
This reward measures how well the selected frames cover the entire video in feature space: for each frame $v_i \in V$, we keep only the distance to its closest selected frame in $\hat{V}^{f[-1]}$ and average these distances over all frames. When the selected frames are placed near the implicit cluster centers of $V$, most frames are close to at least one selection, the average distance is small, and $R_{\mathrm{rep}}$ stays close to $1$~\cite{zhou2018deep}. Conversely, if the policy selects outliers or redundant frames, many frames remain far from any selection, the average distance grows and the exponential term sharply reduces the reward, pushing the policy toward a compact set of prototypical frames that best represent the video.

\paragraph{Curiosity Reward.}
To balance exploration and the tendency to overuse the visual operations, we incorporate a curiosity reward~\cite{su2025pixelreasoner}:
\begin{align}
R_{\text{cur}}(\hat{V}_i)
&= 
\alpha \left( 
H - \frac{1}{K}\sum_{j=1}^{K} \mathbb{I}\big[|\hat{V}_j| > 0\big]
\right)_+ \cdot \mathbb{I}\big[|\hat{V}_i| > 0\big]
\notag\\
&\quad
-\;\beta \left(\,|\hat{V}_i| - N\,\right)_+~,
\end{align}
where $K$ is the number of rollouts, $(\cdot)_+$ is the ReLU function, $\mathbb{I}[\cdot]$ is the indicator function, $\alpha$ and $\beta$ are coeficients, and $H$ is the threshold. The first term becomes positive only when the overall fraction of rollouts that invoke the visual operation falls below $H$, and the factor $\mathbb{I}[|\hat{V}_i| > 0]$ ensures that only rollouts that actually call the tool receive this bonus. This encourages the policy to explore visual tool calls when they are globally under-utilized, rather than collapsing to a purely text-only strategy. In contrast, the second term is activated only when $|\hat{V}_i|$ exceeds $N$, imposing a linear penalty on excessive calls and preventing the policy from over-relying on the visual operation. Together, these two terms guide the policy toward a regime where the tool is used when beneficial, but neither ignored nor abused.

\subsection{Training Framework}

As illustrated in \Cref{fig:training_framework}, we train Video-R4 with a four-stage rumination training framework: Deliberate Rumination Practice (DRP) SFT, a first GRPO-based RL stage, Compositional Rumination Practice (CRP) SFT, and a second RL stage.

% \paragraph{Deliberate Rumination Practice (DRP).}
% We start by teaching the model to master each atomic visual operation in isolation.
% In this Deliberate Rumination Practice SFT (DRP-SFT) stage, each training sample exposes only one type of rumination, either cropping on a single frame or clipping on a video.
% For image-centric samples, the input is one frame, and the model is allowed to crop regions repeatedly, zoom in, and reason over the re-encoded crops.
% For video-centric samples, the input is an ordered list of frames, and the model is allowed to select short consecutive clips from the original sequence and reason over the selected frames.
% DRP-SFT uses about 7k trajectories, including 5k image-based cropping trajectories and 2k video-based clipping trajectories.
% During DRP-SFT, we minimize token-level cross-entropy over both natural language and operation arguments. Hence, the model learns when to invoke cropping or clipping under a single available tool per trajectory and how to propose spatial or temporal regions conditioned on its current rumination state.
% This yields a DRP-initialized checkpoint with robust but non-compositional cropping and clipping skills.
\paragraph{Deliberate Rumination Practice (DRP).}
The first stage focuses on learning each atomic visual operation in isolation. In Deliberate Rumination Practice SFT (DRP-SFT), every training trajectory exposes only one type of rumination: either cropping over a single frame or clipping over a video. For image-centric trajectories, the input consists of one frame, and the model can repeatedly crop regions, zoom in, and reason over the re-encoded crops. For video-centric trajectories, the input is an ordered list of frames, and the model can select short consecutive clips from the sequence and reason over the selected frames. DRP-SFT uses about 7k trajectories, including 5k image-based cropping and 2k video-based clipping examples. We minimize token-level cross-entropy on both natural-language tokens and operation arguments, so the model learns when to call cropping or clipping (under a single available tool per trajectory) and how to propose spatial or temporal regions conditioned on its current rumination state. This stage produces a DRP-initialized checkpoint with strong but still non-compositional cropping and clipping skills.

\paragraph{Compositional Rumination Practice (CRP).}
The second stage teaches the model to interleave cropping and clipping within a single reasoning trajectory. In Compositional Rumination Practice SFT (CRP-SFT), we fine-tune from the DRP-SFT checkpoint on 10k trajectories from Video-R4-CoT-17k, where both atomic operations are available. These trajectories exhibit typical patterns such as first clipping to locate a relevant segment, then cropping on key frames to read fine-grained text, and finally clipping again to verify an earlier hypothesis. Compared with DRP-SFT, CRP-SFT shifts the objective from mastering individual tools to learning longer read–ground–verify procedures, where the model must choose which operation to invoke, how many times to use it, and how to schedule and chain multiple crops and clips over time.

% We then move to Compositional Rumination Practice SFT (CRP-SFT), where the goal is to teach the model to combine cropping and clipping within a single reasoning trajectory.
% To this end, we use 10k trajectories from our Video-R4-CoT-17k in which both atomic operations are available and often interleaved.
% These trajectories expose patterns such as first clipping to locate a relevant segment, then cropping on key frames to read fine-grained text, and finally clipping again to verify an earlier hypothesis.
% Starting from the DRP-SFT checkpoint, we fine-tune on 10k such compositional trajectories in CRP-SFT.
% Compared with DRP-SFT, CRP-SFT shifts the focus from mastering individual tools to learning longer read-ground-verify procedures, where the model must decide which operation to use, how many times to call it, and how to schedule and chain multiple crops and clips over time.

\begin{table*}[!ht]
\centering
\caption{Performance comparison on the testset of the M4-ViteVQA dataset. Best non-human scores in \textbf{bold}. The parameters of the LMM-based models are 7B or 8B. All the RL-based models compared are based on Qwen2.5-VL. The human performances are from \cite{zhao2022vitevqa}.}
\label{tab:m4_vitevqa_main}
\resizebox{\textwidth}{!}{
\begin{tabular}{lccccccccc}
\toprule
\multirow{2}{*}{\makecell[l]{\textbf{Models}}} &
\multirow{2}{*}{\makecell[c]{\textbf{LMM-}\\\textbf{Based}}} &
\multirow{2}{*}{\makecell[c]{\textbf{Visual-}\\\textbf{Grounded}}} & % swapped
\multirow{2}{*}{\makecell[c]{\textbf{RL}-\\\textbf{Based}}} & % swapped
\multicolumn{2}{c}{\makecell[c]{\textbf{Task 1 - Split 1}}} &
\multicolumn{2}{c}{\makecell[c]{\textbf{Task 1 - Split 2}}} &
\multicolumn{2}{c}{\textbf{Task 2}} \\
\cmidrule(lr){5-6}\cmidrule(lr){7-8}\cmidrule(lr){9-10}
& & & & Acc. (\%) & ANLS (\%) & Acc. (\%) & ANLS (\%) & Acc. (\%) & ANLS (\%) \\
\midrule
JustAsk~\cite{yang2021justask}           & \xmark & \xmark & \xmark & 10.05 & 14.10 & 5.47 & 8.60 & 3.60 & 6.70 \\
All-in-one-B~\cite{wang2023allinoneb}      & \xmark & \xmark & \xmark & 10.87 & 14.80 & 5.66 & 7.80 & 3.28 & 4.60 \\
Video-LLaVA-7B~\cite{lin2024videollava}       & \cmark & \xmark & \xmark & 15.43 & 17.15 & 11.19 & 12.02 & 9.38 & 11.80 \\
T5-ViteVQA~\cite{zhao2022vitevqa}        & \xmark & \xmark & \xmark & 22.17 & 29.10 & 16.68 & 23.80 & 9.29 & 13.60 \\
VideoLLaMA2-7B~\cite{cheng2024videollama2}       & \cmark & \xmark & \xmark & 20.76 & 23.55 & 18.33 & 20.45 & 16.54 & 21.08 \\
Qwen2-VL-7B~\cite{wang2024qwen2}          & \cmark & \xmark & \xmark & 35.22 & 45.84 & 27.25 & 38.45 & 21.23 & 28.79 \\
TEA-L~\cite{zhang2025trackandanswer}             & \xmark & \cmark & \xmark & 34.78 & 43.71 & 28.43 & 38.13 & 18.83 & 28.90 \\
NVILA-8B~\cite{liu2025nvila}             & \cmark & \xmark & \xmark & 37.73 & 47.23 & 30.10 & 41.52 & 22.89 & 30.34 \\
GAT-L~\cite{zhang2025gatherandtrace}             & \xmark & \cmark & \xmark & 38.30 & 48.23 & 30.90 & 41.81 & 22.13 & 30.75 \\
Video-R1-7B~\cite{feng2025videor1}          & \cmark & \xmark & \cmark & 37.10 & 48.25 & 33.67 & 44.94 & 43.16 & 53.37 \\
Qwen2.5-VL~\cite{bai2025qwen25}   & \cmark & \xmark & \xmark   & 26.53 & 44.91 & 24.34 & 39.60 & 32.81 & 50.82 \\
Pixel-Reasoner~\cite{su2025pixelreasoner}    & \cmark & \cmark & \cmark & 52.91 & 61.44 & 48.88 & 58.23 & 58.97 & 65.32 \\
\rowcolor{blue!8}\textbf{Video-R4-7B (ours)}   & \cmark & \cmark & \cmark & \textbf{56.17} & \textbf{65.22} & \textbf{52.69} & \textbf{61.89} & \textbf{64.21} & \textbf{69.99} \\
\midrule
Human & -- & -- & -- & 85.27 & 89.30 & 78.41 & 82.80 & 82.26 & 85.10 \\
\bottomrule
\end{tabular}
}
% \vspace{-1em}
\end{table*} 

\paragraph{Multi-stage Rumination Learning.}
Beyond supervised chain-of-thought trajectories, we further refine rumination behavior using GRPO-based RL with the reward $R'$ defined in \Cref{sec:rewards}. We take the Video-R4-RL-30k collection from our data curation pipeline and split it into two subsets of 15k trajectories. Our final schedule is
$\text{DRP-SFT} \rightarrow \text{RL}_{\text{d}} \rightarrow \text{CRP-SFT} \rightarrow \text{RL}_{\text{c}}$.
In the first reinforcement stage $\text{RL}_{\text{d}}$, we start from the DRP-SFT checkpoint and apply GRPO on the first 15k trajectories. This stage encourages the model to explore cropping and clipping under the outcome-based reward $R'$ while staying close to the deliberate single-tool rumination patterns learned in DRP. We then run CRP-SFT on the 10k compositional trajectories so that longer mixed-operation strategies are distilled back into the policy. Finally, the second reinforcement stage $\text{RL}_{\text{c}}$ initializes from the CRP-SFT checkpoint and optimizes on the remaining 15k trajectories, sharpening decisions about when to stop, when to re-zoom, and how aggressively to explore alternative clips and crops.

\section{Experiments}

\subsection{Experiment Setups}

\paragraph{Benchmarks.} We experiment with the text-rich video reasoning on the testset of the M4-ViteVQA~\cite{zhao2022vitevqa} dataset. Models are also tested on three commonly-used general video QA benchmarks: MVBench~\cite{li2024mvbench}, Video-MME~\cite{fu2025videomme}, and Video-MMMU~\cite{hu2025videommmu}. The generalization ability is evaluated on the multi-page document QA benchmark MP-DocVQA~\cite{tito2023hierarchical} and slides QA benchmark SlidesVQA~\cite{tanaka2023slidevqa}.

\paragraph{Evaluation Metrics.}
For text-rich video QA and multi-page document QA, we use accuracy (exact match, EM)~\cite{rajpurkar2016squad,yang2018hotpotqa} and Average Normalized Levenshtein Similarity (ANLS)~\cite{biten2019scene}.
For general video QA, we use accuracy.
For slides QA, we use EM~\cite{rajpurkar2016squad,yang2018hotpotqa} and Macro-averaged F1 score. The details about EM, ANLS, and (Macro-averaged) F1 can be found in the appendix (\Cref{app:sec:eval_metrics}).

\paragraph{Baselines.}
We compare Video-R4-7B against three groups of representative methods:
(1) \emph{Conventional video QA and long-context language models} that do not rely on multimodal LMMs, including JustAsk~\cite{yang2021justask}, All-in-one-B~\cite{wang2023allinoneb}, T5-ViteVQA~\cite{zhao2022vitevqa}, and long-sequence Transformers such as Longformer~\cite{beltagy2020longformer} and Big-Bird~\cite{zaheer2020big}.
(2) \emph{Instruction-tuned video LMMs} without explicit reasoning-by-grounding or RL, e.g., Video-LLaVA~\cite{lin2024videollava}, VideoLLaMA2~\cite{cheng2024videollama2}, Qwen2-VL~\cite{wang2024qwen2}, Qwen2.5-VL~\cite{bai2025qwen25}, and NVILA~\cite{liu2025nvila}; and \emph{visual-grounded} designs tailored for text-rich videos, such as TEA-L~\cite{zhang2025trackandanswer} and GAT-L~\cite{zhang2025gatherandtrace}.
(3) \emph{RL-tuned reasoning LMMs} built on strong LMM backbones, including Video-R1~\cite{feng2025videor1} and Pixel-Reasoner~\cite{su2025pixelreasoner}. We also report human performance from the M4-ViteVQA benchmark~\cite{zhao2022vitevqa}. Unless otherwise noted, LMM-based competitors use 7B/8B backbones, matching our model size for a fair comparison.

\begin{table*}[!ht]
\centering
\caption{Ablation results of Video-R4 training framework, tested on the testset of M4-ViteVQA dataset. Best scores in \textbf{bold} and second-best in \underline{underline}.}
\vspace{-0.2cm}
\label{tab:m4_vitevqa_ablation}
\resizebox{\textwidth}{!}{
\begin{tabular}{lcccccc}
\toprule
\multirow{2}{*}{\makecell[l]{\textbf{Training Framework}}} &
\multicolumn{2}{c}{\makecell[c]{\textbf{Task 1 - Split 1}}} &
\multicolumn{2}{c}{\makecell[c]{\textbf{Task 1 - Split 2}}} &
\multicolumn{2}{c}{\textbf{Task 2}} \\
\cmidrule(lr){2-3}\cmidrule(lr){4-5}\cmidrule(lr){6-7}
& Acc. (\%) & ANLS (\%) & Acc. (\%) & ANLS (\%) & Acc. (\%) & ANLS (\%) \\
\midrule
\rowcolor{blue!8}DRP-SFT $\rightarrow$ RL$_{\text{d}}$ $\rightarrow$ CRP-SFT $\rightarrow$ RL$_{\text{c}}$ (full) & \textbf{56.17} & \underline{65.22} & \textbf{52.69} & \underline{61.89} & \textbf{64.21} & 69.99 \\
DRP-SFT $\rightarrow$ RL$_{\text{d}}$ $\rightarrow$ CRP-SFT $\rightarrow$ RL$_{\text{c}}$ (w/o $R_{\text{rep}}$)     & 55.70 & 65.04 & \underline{52.54} & 61.86 & \underline{62.65} & \textbf{71.08} \\
DRP-SFT $\rightarrow$ RL$_{\text{d}}$ $\rightarrow$ CRP-SFT $\rightarrow$ RL$_{\text{c}}$ (w/o $R_{\text{div}}$)     & 55.56 & \textbf{65.24} & 52.26 & \textbf{62.13} & 62.41 & \underline{70.11} \\
DRP-SFT $\rightarrow$ RL$_{\text{d}}$ $\rightarrow$ CRP-SFT $\rightarrow$ RL$_{\text{c}}$ (w/o $R_{\text{cur}}$)     & 54.35 & 63.92 & 50.90 & 61.10 & 61.92 & 69.20 \\
DRP-SFT $\rightarrow$ RL$_{\text{d}}$ $\rightarrow$ CRP-SFT $\rightarrow$ RL$_{\text{c}}$ (w/o $R_{\text{div}}$, $R_{\text{rep}}$)      & \underline{55.73} & 64.99 & 52.02 & 61.35 & 62.24 & 68.38 \\
DRP-SFT $\rightarrow$  CRP-SFT $\rightarrow$ RL$_{\text{c}}$                  & 54.98 & 63.74 & 51.50 & 60.80 & 60.44 & 68.59 \\
CRP-SFT $\rightarrow$ RL$_{\text{c}}$                       & 54.23 & 63.75 & 51.26 & 60.39 & 61.43 & 68.28 \\
DRP-SFT $\rightarrow$ RL$_{\text{d}}$ $\rightarrow$ CRP-SFT                 & 50.08 & 64.15 & 46.17 & 60.67 & 56.27 & 68.81 \\
CRP-SFT                            & 46.76 & 62.67 & 40.47 & 59.68 & 49.47 & 66.33 \\
DRP-SFT $\rightarrow$ CRP-SFT                      & 44.58 & 63.09 & 42.23 & 60.58 & 51.92 & 69.07 \\
DRP-SFT                            & 32.80 & 57.92 & 32.07 & 54.74 & 33.74 & 63.70 \\
Base Model (Qwen2.5-VL-7B-Instruct)      & 26.53 & 44.91 & 24.34 & 39.60 & 32.81 & 50.82 \\
\bottomrule
\end{tabular}
}
\end{table*}
% \vspace{-0.2cm}

\subsection{Main Results}

\paragraph{Text-Rich Video QA.}
\Cref{tab:m4_vitevqa_main} summarizes results on the M4-ViteVQA testset~\cite{zhao2022vitevqa}. Video-R4-7B establishes a new state of the art among non-human systems across all three evaluation splits. The largest margin appears on Task 2, where Video-R4-7B reaches 64.21 Acc against 43.16 for Video-R1~\cite{feng2025videor1}. The Improvements are consistent in both accuracy and ANLS metrics. Beyond the aggregate scores, we also observe that allowing the model to execute a longer sequence of visual operations at inference monotonically improves accuracy. When the inference allows deeper visual rumination, performance scales up. This aligns with the broader test-time scaling phenomenon and indicates that longer visual rumination increases the chance of finding and verifying small text cues rather than relying on a single pass.

\begin{myboxi}\textbf{Finding 1:} 
% Fixed-format CoT supervision enables imitation of reasoning patterns. 
Allowing longer rumination and more pixel-grounded steps consistently boosts accuracy, demonstrating test-time scaling effect.
\end{myboxi}
% \vspace{-2.5em}

\paragraph{Ablation Study.}
\Cref{tab:m4_vitevqa_ablation} compares training recipes. The full schedule DRP-SFT $\rightarrow$ RL\textsubscript{d} $\rightarrow$ CRP-SFT $\rightarrow$ RL\textsubscript{c} achieves the best end performance and also converges faster early on than direct CRP-SFT or DRP-SFT followed by CRP-SFT. 
The benefit remains even when later losses become similar. On Task 2, Acc moves from 51.92 for DRP-SFT $\rightarrow$ CRP-SFT and 61.43 for CRP-SFT $\rightarrow$ RL\textsubscript{c} to 64.21 with the full recipe.
During RL the policy develops a marked preference for cropping over clipping. Cropping isolates and enlarges a single informative frame, which reduces redundancy and helps read fine text, consistent with how humans pause and zoom when analyzing videos. Reward ablations indicate a trade-off between repetition and diversity controls, which shape the balance between careful reading and broad exploration.
\begin{myboxi}\textbf{Finding 2:} 
% Fixed-format CoT supervision enables imitation of reasoning patterns. 
The \emph{DRP $\rightarrow$ RL $\rightarrow$ CRP $\rightarrow$ RL} schedule yields the best performance, indicating that atomic, first then compositional learning, interleaved with RL, is most effective.
\end{myboxi}

\subsection{Generalization Experiments}

\paragraph{General Video QA.}
Without dataset-specific tuning, Video-R4-7B transfers competitively to general video QA as summarized in \Cref{tab:three-in-one}. It is near the top on MVBench~\cite{li2024mvbench} and Video-MME~\cite{fu2025videomme} and sets a new best 52.2 on Video-MMMU~\cite{hu2025videommmu}. Video-MMMU contains many educational/lecture videos that are intrinsically text-rich, and the read–ground–verify routine learned on M4-ViteVQA appears to be well-aligned with these data.
\begin{myboxi}\textbf{Finding 3:} 
% Fixed-format CoT supervision enables imitation of reasoning patterns. 
RL encourages a preference for cropping over clipping, as zooming provides more informative and less redundant evidence, mirroring how humans pause and inspect frames.
\end{myboxi}
\paragraph{Multi-page Document \& Slides QA.}
After training on text-rich video QA, Video-R4-7B transfers to long document understanding with no additional tuning.
On MP-DocVQA, the zero-shot result of Video-R4-7B is 53.21 Acc and 62.22 ANLS, surpassing both the zero-shot and trained Hi-VT5 variants~\cite{tito2023hierarchical}. On the testset of SlidesVQA, Video-R4-7B reaches 43.0 EM and 52.2 F1 versus 33.5 and 41.7 for M3D~\cite{tanaka2023slidevqa} as shown in \Cref{tab:three-in-one}. These demonstrate that once trained to locate, read, and verify dispersed textual evidence over time, the model can reuse the same procedure across pages and slides with minimal friction.
\begin{myboxi}\textbf{Finding 4:} Training on text-rich videos transfers well to multi-page documents, slides, and general video QA, with strong gains on the text-heavy Video-MMMU benchmark.
\end{myboxi}

% in document
\begin{table*}[!ht]
\centering
\caption{Fine-tuning on Video-R4-CoT-17k and Video-R4-RL-30k, Video-R4 demonstrates strong generalization capabilities, effectively handling not only general video QA but also multi-page document QA and slides QA, without the need for further dataset-specific training.}
\label{tab:three-in-one}

% ---------- LEFT COLUMN (Table 3 over Table 4) ----------
\begin{minipage}[t]{0.47\textwidth}
\vspace{0pt}

% --- Table 3 ---
\subcaptionbox{Results on general video QA benchmarks.\label{tab:video-benchmarks}}{%
\resizebox{\linewidth}{!}{%
\begin{tabular}{lccc}
\toprule
\textbf{Models} & \textbf{MVBench} & \textbf{Video-MME} & \textbf{Video-MMMU} \\
\midrule
Video-LLaVA-7B~\cite{lin2024videollava}        & 42.9 & 39.9 & --   \\
VideoLLaMA2-7B~\cite{cheng2024videollama2}     & {54.6} & 46.6 & --   \\
Qwen2.5-VL-7B~\cite{bai2025qwen25}             & 57.4 & 53.1 & 47.8 \\
Video-R1-7B~\cite{feng2025videor1}             & 62.7 & \textbf{57.4} & \underline{49.8} \\
Pixel-Reasoner~\cite{su2025pixelreasoner}   & \textbf{65.4} & \underline{54.6} & {47.7} \\
\rowcolor{blue!8}\textbf{Video-R4-7B (ours)}  & \underline{64.5} & 54.5 & \textbf{52.2} \\
\bottomrule
\end{tabular}%
}
} % end subcaption

\vspace{0.9em}

% --- Table 4 ---
\subcaptionbox{Results on the validation set of the MP-DocVQA dataset.\label{tab:simplified}}{%
\resizebox{\linewidth}{!}{%
\begin{tabular}{l c c c}
\toprule
\textbf{Models} & \textbf{Zero-Shot} & \textbf{Acc. (\%)} & \textbf{ANLS (\%)} \\
\midrule
% BERT~\cite{devlin2019bert}      & \xmark & 27.41 & 41.83 \\
LayoutLMv3~\cite{huang2022layoutlmv3}      & \xmark & 38.47 & 45.38 \\
Big-Bird~\cite{zaheer2020big}      & \xmark & 41.06 & 49.29 \\
Hi-VT5 (w/o train)~\cite{tito2023hierarchical}  & \cmark & 42.10 & 58.64 \\
% T5~\cite{raffel2020exploring}      & \xmark & 41.80 & 50.50 \\
Longformer~\cite{beltagy2020longformer}    & \xmark & 43.91 & 52.87 \\
Hi-VT5~\cite{tito2023hierarchical}  & \xmark & 48.28 & 62.01 \\
\rowcolor{blue!8}\textbf{Video-R4-7B (ours)}  & \cmark & \textbf{53.21} & \textbf{62.22} \\
\bottomrule
\end{tabular}%
} % end resize
} % end subcaption

\end{minipage}
\hfill
% ---------- RIGHT COLUMN (Table 5) ----------
\begin{minipage}[t]{0.475\textwidth}
\vspace{0pt}

\subcaptionbox{Results on the test set of SlidesVQA.\label{tab:results}}{%
\resizebox{\linewidth}{!}{%
\begin{tabular}{l c c c c c}
\toprule
\multirow{2}{*}{\textbf{Models}} & \multirow{2}{*}{\makecell{\textbf{Zero-}\\\textbf{Shot}}} & \multicolumn{2}{c}{\textbf{Dev}} & \multicolumn{2}{c}{\textbf{Test}} \\
\cmidrule(lr){3-4}\cmidrule(lr){5-6}
 &  & EM & F1 & EM & F1 \\
\midrule
Q-only~\cite{tanaka2023slidevqa}                     & \cmark &  9.4 & 11.4 & 10.7 & 13.5 \\
UniVL~\cite{luo2020univl}                      & \cmark &  8.8 & 12.1 & 10.6 & 14.1 \\
PreasM~\cite{yoran2022turning}                     & \xmark & 36.3 & 41.9 & 30.7 & 38.2 \\
T5~\cite{raffel2020exploring}                         & \xmark & 35.2 & 41.3 & 29.3 & 37.9 \\
T5 + $z^{\text{lay}}$~\cite{raffel2020exploring}      & \xmark & 36.9 & 43.2 & 31.0 & 39.7 \\
LayoutT5~\cite{tanaka2021visualmrc}                   & \xmark & 38.9 & 44.8 & 31.7 & 39.9 \\
LayoutLMv2~\cite{tu2020select}                 & \xmark & 26.5 & 33.4 & 21.4 & 29.3 \\
FiD~\cite{izacard2021leveraging}                        & \xmark & 37.6 & 42.9 & 30.4 & 38.9 \\
FiD + $z^{\text{lay}}$~\cite{izacard2021leveraging}     & \xmark & 38.1 & 43.3 & 30.6 & 38.9 \\
M3D~\cite{tanaka2023slidevqa}                        & \xmark & {41.3} & {47.1} & {33.5} & {41.7} \\
\rowcolor{blue!8}\textbf{Video-R4-7B (ours)} & \cmark & \textbf{49.5} & \textbf{56.0} & \textbf{43.0} & \textbf{52.2} \\
\midrule
Human                              & --     & --   & --   & 89.8 & 93.0 \\
\bottomrule
\end{tabular}%
} % end resize
} % end subcaption

\end{minipage}
\end{table*}
% \vspace{-0.3cm}
\section{Related Work}
\paragraph{Text-Rich Video Understanding.}
Early work on text-rich visual understanding primarily studied single images, where TextVQA systems integrate OCR, layout cues, and semantic reasoning to read scene text~\cite{hu2020iterative,Bi_2021_ICCV,zhou2023exploring,guo2024benchmarking,zhou2024graph}. TextVideoQA extends this problem to dynamic scenes, requiring models to track temporal changes and transient text signals~\cite{zhao2022vitevqa,tom2023reading,jahagirdar2023watching,zhou2025egotextvqa}. M4-ViteVQA~\cite{zhao2022vitevqa} provides the first large-scale benchmark, while later datasets such as RoadTextVQA and NewsVideoQA~\cite{tom2023reading,jahagirdar2023watching} study domain-specific settings. Beyond general video LMMs, several architectures explicitly optimize grounding over text regions: TEA-L and GAT-L introduce tracking and graph reasoning over OCR boxes~\cite{zhang2025trackandanswer,zhang2025gatherandtrace}, and Pixel-Reasoner employs pixel-level cropping actions for fine-grained evidence acquisition~\cite{su2025pixelreasoner}. Related tasks such as multi-page document QA and slides QA tackle long-range textual grounding using hierarchical or layout-aware encoders~\cite{tito2023hierarchical,tanaka2023slidevqa,beltagy2020longformer,zaheer2020big,huang2022layoutlmv3}. In contrast, Video-R4 builds on a generic video LMM and learns explicit spatio-temporal operations through RL.

% \vspace{-0.1cm}
\paragraph{Video Understanding with LMMs.}
Visual instruction tuning has established a strong foundation for aligning image--language models with general-purpose language backbones~\cite{zhu2023minigpt,liu2023llava,chen2024far,liu2023llava,li2023blip,tang2025videolmmposttrainingdeepdive,zhang2025diversifying}. Building upon this paradigm, video LMMs extend 2D visual alignment to temporal sequences. Early methods rely on sparse frame sampling and lightweight temporal fusion~\cite{li2023blip,liu2024oryx}, while recent systems strengthen spatial--temporal modeling, token compression, and long-context handling. Video-LLaVA~\cite{lin2024videollava} and VideoLLaMA2~\cite{cheng2024videollama2} incorporate temporal attention and audio streams; Qwen2-VL and Qwen2.5-VL adopt dynamic-resolution inputs for higher-fidelity video perception~\cite{wang2024qwen2,bai2025qwen25}; NVILA further scales visual backbones for video LMMs~\cite{liu2025nvila}. For long or streaming videos, works such as LongVA~\cite{zhang2024long}, Video-XL and successors~\cite{shu2024video,liu2025video}, VideoChat/VideoChatFlash~\cite{li2023videochat,li2024videochatf}, InternVideo2~\cite{wang2024internvideo2}, and MVBench/Video-MME~\cite{li2024mvbench,fu2025videomme} explore compression, memory, and benchmarking. Most existing approaches~\cite{Bi_2024}, however, still treat video understanding as a single-pass perception task over fixed frames, without modeling iterative zoom-and-check behaviors on text-rich regions. Video-R4 instead performs closed-loop visual rumination, enabling deliberate evidence gathering and multi-step grounding.

% \vspace{-0.1cm}
\paragraph{Large Multimodal Model Reasoning.}
LLM reasoning has progressed from chain-of-thought supervision~\cite{wei2022chain,zhou2022least,zhang2023multimodal,bi2025reasoningmatterssurveyadvancements,bi2025diagnosingvisualreasoningchallenges} to outcome-driven RL, where rule-based rewards and GRPO enable strong long-form reasoning without dense annotations~\cite{jaech2024openai,guo2025deepseek,chen2025towards,li2025system}. This paradigm has inspired advances in text reasoning~\cite{yuan2024advancing,zhang2025critique,yuan2025mme} and domain-specific R1-style models in math, finance, and medicine~\cite{shao2024deepseekmath,liu2025fin,lai2025med,Zhang_2025_ICCV}. For vision–language models, recent work applies verifiable visual rewards to teach multi-step perceptual reasoning~\cite{huang2025vision,r1vl,visualrft,segzero,deng2025openvlthinker,fan2025sophiavl,zhou2025r1,drift,bi2025verifybenchmarkvisualexplanation,sun2025latent}. Closer to our task, \cite{videor1,wang2025timezero,r1omni} explore temporal grounding in videos using GRPO objectives.
Video-R4 differs by reasoning with a structured tool interface and a tailored reward scheme. Our diversity/representativeness rewards shape spatial–temporal coverage, while a curiosity reward regulates tool frequency. Combined with the DRP/CRP rumination curriculum, this encourages human-like zoom-and-read behaviors crucial for recovering small, dispersed text in long videos and multi-page documents.

\section{Conclusion}
We presented Video-R4, a video reasoning agent that acquires evidence through iterative visual rumination. By decomposing video understanding into frame selection, spatial zooming, and re-encoding cycles, Video-R4 overcomes the limitations of single-pass perception and enables reliable grounding on text-rich, fine-grained visual cues. To support this paradigm, we curated Video-R4-CoT-17k and Video-R4-RL-30k, providing the first executable supervision for multi-step video rumination. Our multi-stage training strategy, combining supervised trajectories with GRPO-based reinforcement learning, proves essential for stabilizing and improving rumination behavior. Empirical results show that Video-R4 achieves state-of-the-art performance on M4-ViteVQA and generalizes to broader multimodal reasoning tasks. We believe rumination-based LMMs can extend to longer videos, multimodal evidence fusion, and more open-ended reasoning, pushing LMMs toward robust, human-like video understanding.

\paragraph{Acknowledgements.}
This work was supported by Sony Group Corporation. We would like to thank Sayaka Nakamura and Jerry Jun Yokono for their insightful discussion.
{
    \small
    \bibliographystyle{ieeenat_fullname}
    \bibliography{main}
}

\clearpage
\setcounter{page}{1}
\maketitlesupplementary

\section{Limitations}
Despite these results, Video-R4 still has several limitations. First, the data curation pipeline relies on pre-extracted OCR results and object detections, so recognition errors or missing text can directly hurt both rumination trajectories and final answers. Second, the current tool interface supports only frame selection and spatial cropping with a bounded trajectory length, which may be insufficient for very long or fast-changing videos that require richer operations (e.g., tracking, retiming, or audio-aware cues). Third, our training data are primarily derived from M4-ViteVQA and a few related text-centric datasets, and experiments are conducted on a 7B backbone, leaving open questions about robustness under more diverse domains and larger model scales. Finally, the GRPO reward combines hand-designed proxies such as diversity, representativeness, and curiosity, which only approximate human notions of faithfulness and interpretability. Future work could relax these assumptions by broader operation types, more diverse optimization methods, and rewards.

\section{Dataset Details}

\begin{figure}[!ht]
    \centering
    \includegraphics[width=\linewidth]{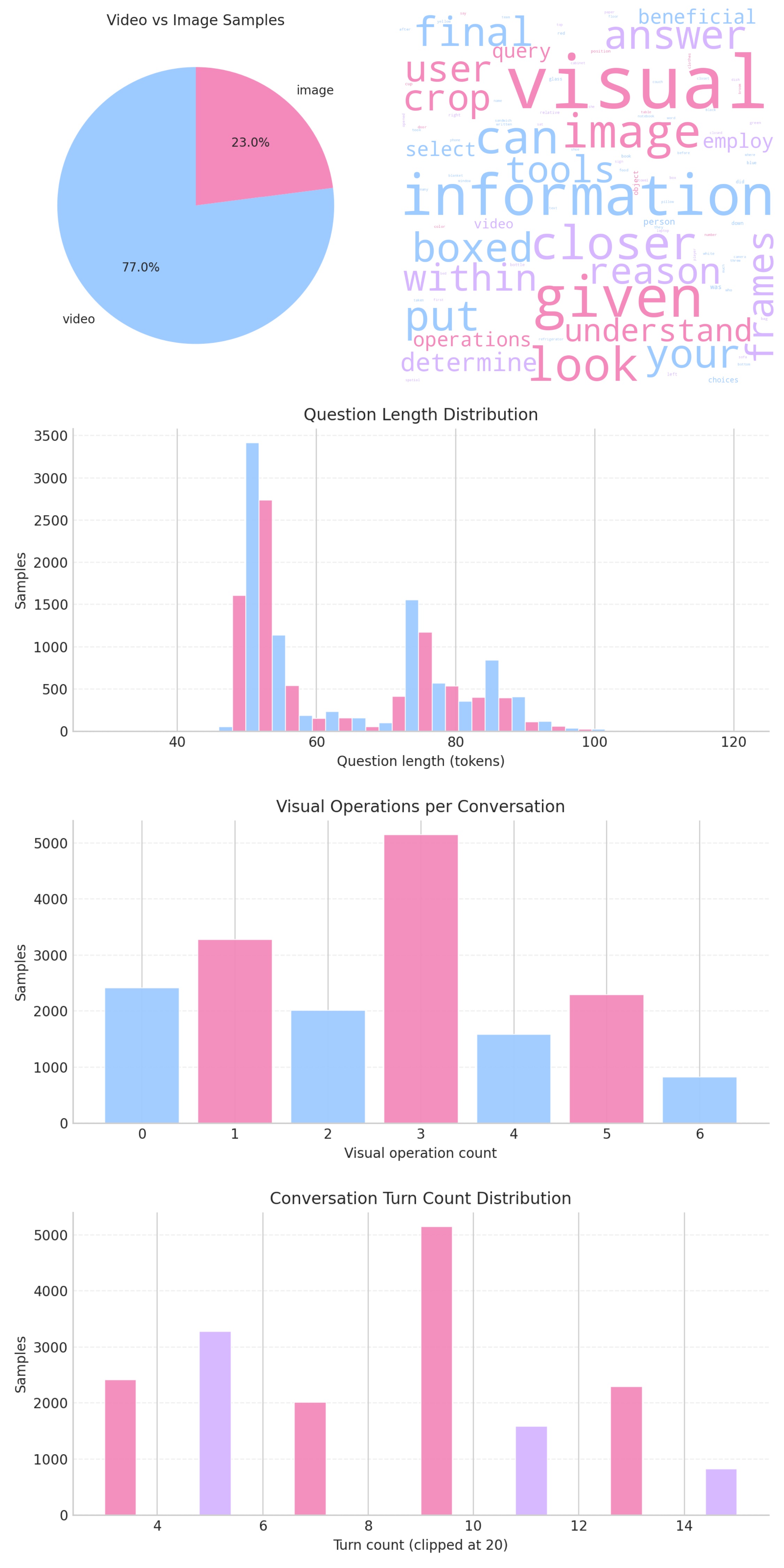}
    \caption{Overall statistics of the {Video-R4-CoT-17k} dataset, including the ratio of video versus image samples, word cloud of frequently appearing terms, question length distribution, distribution of visual operation counts per sample, and conversation turn count distribution.}
    \label{fig:cot_stat}
    \vspace{-1em}
\end{figure}

\begin{figure}[!ht]
    \centering
    \includegraphics[width=\linewidth]{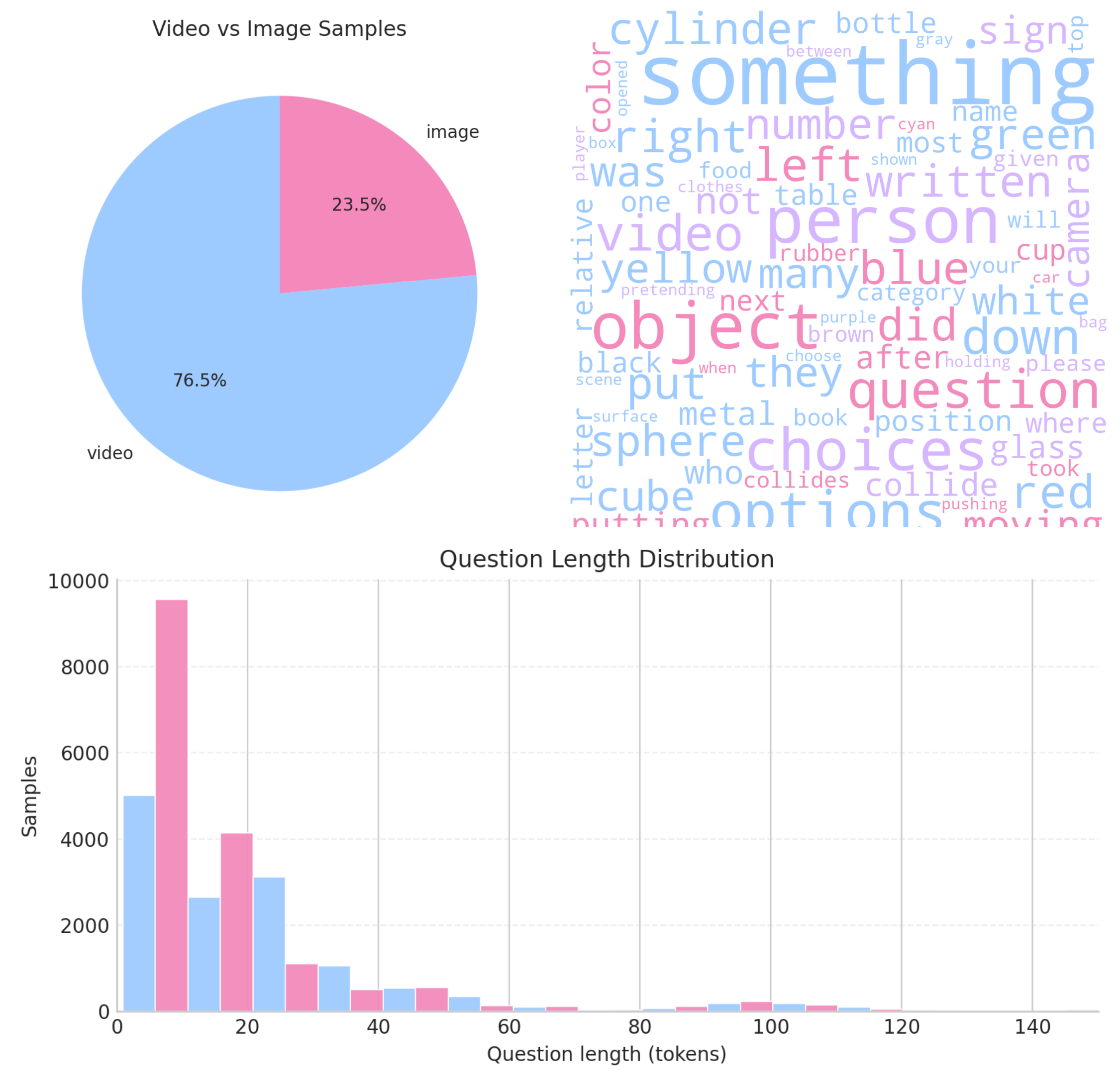}
    \caption{Overall statistics of the {Video-R4-RL-30k} dataset.}
    \label{fig:rl_stat}
    \vspace{-1em}
\end{figure}

\paragraph{Dataset Statistics.}
\Cref{fig:cot_stat} presents the overall statistics of {Video-R4-CoT-17k}. 
The dataset is predominantly video-based, with images forming a smaller subset. 
The word cloud highlights frequent reasoning-related expressions such as ``visual'', ``information'', and various operation-oriented verbs. 
The question length distribution centers on medium-length prompts, while the plots of visual operation counts and conversation turns show that CoT trajectories typically require several visual operations and involve multi-round interactions.
\Cref{fig:rl_stat} summarizes the statistics of {Video-R4-RL-30k}. 
The corresponding word cloud shows a more object-focused vocabulary (e.g., ``object'', ``person'', ``left'', ``color''), consistent with the concise, direct style characteristic of RL-refined queries.

\paragraph{Rule-Based Evidence Matching.}
The Rule-Based Evidence Matching algorithm is shown in \Cref{alg:abstract}.
For each training instance $q$, we denote by $q^{\mathrm{text}}$ the question text and by $q^{\mathrm{ans}}$ the answer expression, which may be a single string or a small set of candidates; the associated video is $v(q)$. Each instance carries two supervisory attributes: the temporal specification $\mathrm{src}_1(q)\in\{\text{Single frame},\text{Multi frame}\}$, indicating whether evidence is restricted to one frame or may span multiple frames, and the modality specification $\mathrm{src}_2(q)\in\{\text{Text},\text{Visual}\}$, indicating whether evidence is primarily textual (OCR) or visual (objects). We write $A_q=\mathrm{tok}(q^{\mathrm{ans}})$ and $W_q=\mathrm{tok}(q^{\mathrm{text}})$ for the normalized token sets of the answer and question, respectively.
For a video $v$, let $F_v$ be the set of candidate frames considered during evidence mining. For every frame $f\in F_v$, we assume paragraph-level OCR regions $\mathcal{P}_{v,f}$, fine-grained OCR text detections $\mathcal{T}_{v,f}=\{(s_{v,f,i},b^{\mathrm{text}}_{v,f,i})\}_i$ with strings and corresponding boxes, and object detections $\mathcal{O}_{v,f}=\{(\ell_{v,f,k},b^{\mathrm{obj}}_{v,f,k})\}_k$ with discrete labels and boxes. Matching and geometry are treated abstractly via the primitives $\mathrm{score}_{\mathrm{text}}(s,A)\in[0,1]$ for text–answer relevance, $\mathrm{score}_{\mathrm{name}}(n,U)\in[0,1]$ for name–token compatibility, $\mathrm{iou}(b_1,b_2)$ for box overlap, $\mathrm{extend}(b)$ for deterministic enlargement, and $\mathrm{merge}(\mathcal{B})$ for minimal axis-aligned merging of a box set $\mathcal{B}$. The goal is, for each question $q$, to return a subset $\mathcal{R}_q\subseteq F_{v(q)}$ of relevant frames and a per-frame evidence region $B^{\mathrm{ev}}_{q,f}$ obtained by combining textual and, when applicable, object cues. We denote by $b^{\mathrm{text}}_{q,f}$ the best OCR-derived box selected in frame $f$ for question $q$ before paragraph refinement.

\begin{algorithm*}[t]
\caption{Rule-Based Evidence Matching}
\label{alg:abstract}
\begin{algorithmic}[1]
\For{each question $q$}
  \State initialize $\mathcal{R}_q \gets \emptyset$
  \For{each frame $f \in F_{v(q)}$}
     \State find best OCR match $b^{\mathrm{text}}_{q,f}$
           using $\mathrm{score}_{\mathrm{text}}(\cdot, A_q)$
     \If{a match exists} $\mathcal{R}_q \gets \mathcal{R}_q \cup \{f\}$ \EndIf
  \EndFor

  \For{each $f \in \mathcal{R}_q$}
     \State refine $b^{\mathrm{text}}_{q,f}$ by selecting
           $p^\star \in \mathcal{P}_{v,f}$ with maximal $\mathrm{iou}$,
           then set $b^{\mathrm{text}}_{q,f} \gets \mathrm{extend}(p^\star)$
  \EndFor

  \If{$\mathrm{src}_2(q)=\mathrm{Text}$}
     \State choose single or multiple frames according to $\mathrm{src}_1(q)$
     \State output refined text boxes $\{b^{\mathrm{text}}_{q,f}\}$
     \State \textbf{continue}
  \EndIf

  \For{each $f \in \mathcal{R}_q$}
     \State collect object boxes whose names match $A_q \cup W_q$
           via $\mathrm{score}_{\mathrm{name}}$
     \State merge all matched boxes with $b^{\mathrm{text}}_{q,f}$:
           $B^{\mathrm{ev}}_{q,f} \gets \mathrm{merge}(\cdot)$
  \EndFor

  \State select single or multiple frames according to $\mathrm{src}_1(q)$
  \State output $\mathcal{R}_q$ and $\{B^{\mathrm{ev}}_{q,f}\}$
\EndFor
\end{algorithmic}
\end{algorithm*}

\begin{figure*}
    \centering
    \includegraphics[width=\linewidth]{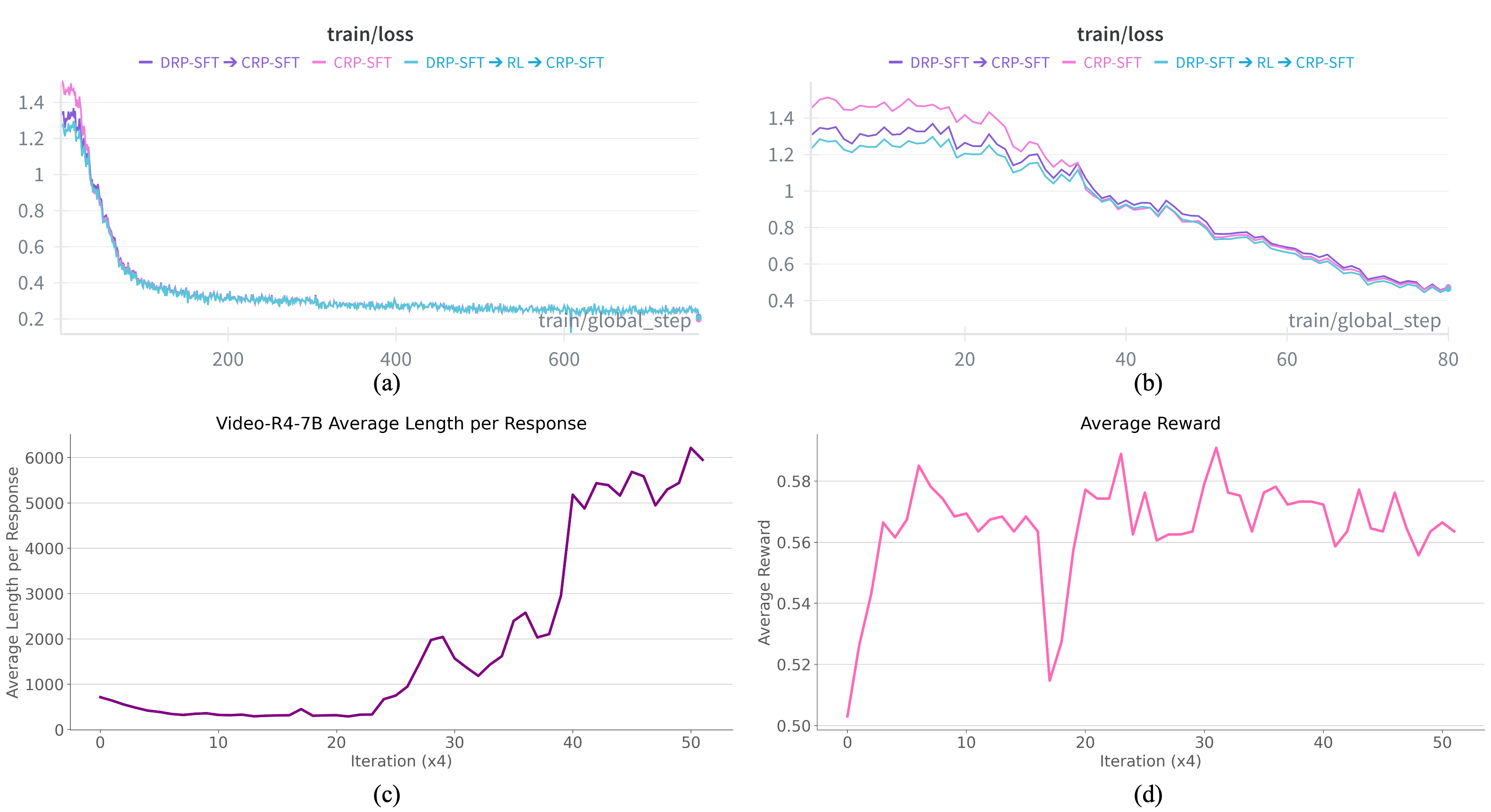}
    \caption{Comparison of training behaviors across fine-tuning strategies. Subfigures (a) and (b) show that models pre-finetuned on DRP-SFT data converge more quickly and achieve lower final loss when training on CRP-SFT, indicating that decomposing visual operations before interleaved training is beneficial. (c) Video-R4-7B progressively increases its response length during RL, suggesting emergent allocation of more thinking time. (d) Correspondingly, the average reward improves and remains stable across iterations.}
    \label{fig:app_line_chart}
\end{figure*}

\paragraph{Template-Based Context Synthesis.}
We construct a set of multi-turn dialogues for each annotation through predefined templates. Each dialogue contains a system message, a question, the path to the input video, and a sequence of turns. The turns follow a chain-of-thought format. Each turn provides an analysis of the visual information obtained through the visual operation applied in the previous step. The first turn, instead, provides an overall description of the input video. The turn then continues with a brief reasoning segment that connects to the next action and ends with a statement describing the next visual operation, where the format of the visual operation follows \cite{su2025pixelreasoner}, with ``$<$tool\_call$>$'' labels to prompt the visual operations, and tool names and parameters are needed for a single function call. For clipping, the parameters are the indices of the selected frames. For cropping, the parameters include a frame index and the bounding box coordinates. The final turn predicts the answer to the question with ``\boxed\{\}'' format. 
The template produces a dialogue that contains several placeholders. These placeholders include the input video caption and a descriptive analysis of the visual observations obtained in each turn. They will be filled in during the following stages.

\paragraph{LMM-Based CoT Synthesis and Refinement.} We use Qwen2.5-VL~\cite{bai2025qwen25} to generate video captions, clip captions, and region captions. The input includes the original video frame sequence for the video captions and the text prompts. For clip and region captions, the original video frame sequence, the clips/regions in the current turn, and the text in the context serve as input. The think processes are then generated, focusing on whether the current visual cues obtained can answer the question sufficiently. Then we replace all the placeholders in the templates to get the multi-turn CoT trajectories. We use GPT-4o~\cite{hurst2024gpt4o} to further refine the trajectories to make them more coherent, natural, and reasonable.

\begin{figure}[!ht]
    \centering
    \includegraphics[width=\linewidth]{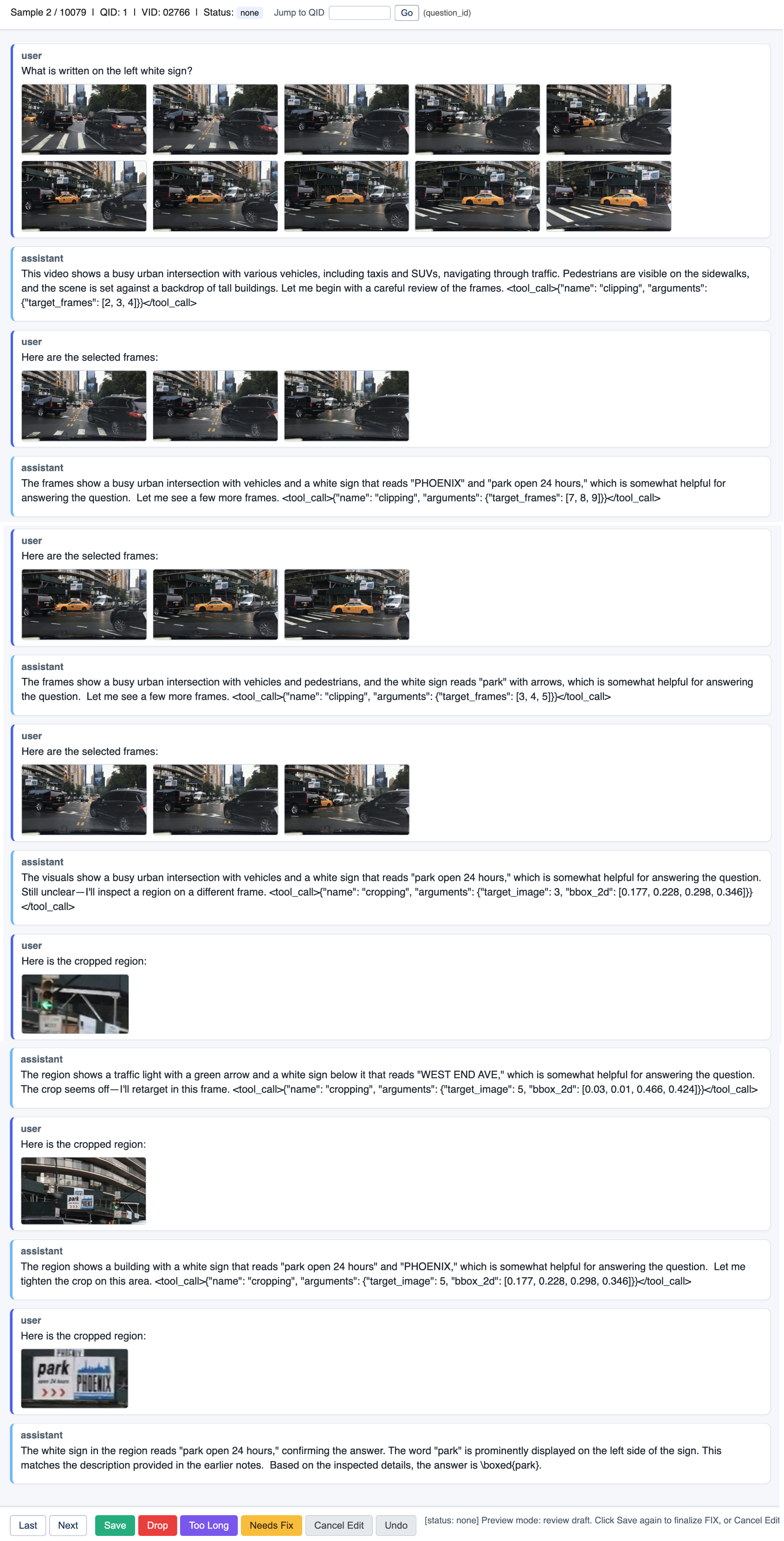}
    \caption{Interface of the quality control tool used to review QA queries and synthesized chain-of-thought trajectories. The tool enables rapid browsing, frame inspection, saving or dropping samples, and in-place editing of both textual and visual reasoning steps to streamline annotation and correction workflows.}
    \label{fig:anno_tool}
    \vspace{-1em}
\end{figure}

\paragraph{Quality Control Tool.}
We develop a quality control tool to quickly review all the QA queries and the corresponding synthesized trajectories. As shown in \Cref{fig:anno_tool}, the tool supports quick browsing, sample saving, dropping functions, and a fixing mode, where human annotators can directly revise the content of the chain-of-thought trajectories, including both text and visual cues.

\section{Evaluation Metrics}
\label{app:sec:eval_metrics}

\paragraph{Average Normalized Levenshtein Similarity (ANLS).}
Exact-match metrics are brittle for text-centric VQA because minor OCR errors can flip a correct rationale into an incorrect string. Therefore, \cite{biten2019scene} proposed ANLS, which turns the normalized Levenshtein distance~\cite{levenshtein1966binary} between a prediction and reference into a similarity score with a cutoff.
Let $o_{q_i}$ be the model's answer for question $q_i$, and $\{a_{ij}\}_{j=1}^{M}$ the set of $M$ ground-truth strings. Denote by $NL(\cdot,\cdot)\!\in[0,1]$ the normalized Levenshtein distance. With a threshold $\tau=0.5$, the per-pair similarity is
\begin{equation}
s(a_{ij}, o_{q_i}) =
\begin{cases}
1 - NL(a_{ij}, o_{q_i}) & \text{if } NL(a_{ij}, o_{q_i}) < \tau, \\
0 & \text{otherwise}.
\end{cases}
\label{eq:anls_score}
\end{equation}
Then take the best match across references for each question and average over $N$ questions:
\begin{equation}
\mathrm{ANLS} = \frac{1}{N} \sum_{i=1}^{N} \left( \max_{j} s(a_{ij}, o_{q_i}) \right).
\label{eq:anls}
\end{equation}
Predictions with edit distance $\ge \tau$ (over half the characters wrong) receive zero credit, while smaller deviations are rewarded proportionally. This softly penalizes OCR noise while still emphasizing exactness.

\paragraph{Exact Match (EM).} This metric quantifies the proportion of predictions that exactly coincide with any of the ground truth answers, thereby providing a strict measure of answer correctness~\cite{rajpurkar2016squad}.

\paragraph{(Macro-averaged) F1 score.} This metric assesses the token-level overlap between a prediction and the ground truth answer by treating both as bags of tokens and computing their F1 score. For each question, the highest F1 score across all ground truth answers is selected, and the final metric is obtained by averaging these maxima over the full set of questions~\cite{rajpurkar2016squad}.

\section{More Evidence to Support our Findings}
\Cref{fig:app_line_chart} (c) demonstrates Video-R4-7B naturally learns to solve reasoning tasks with more thinking time, which is evidence to support our Finding 1. \Cref{fig:app_line_chart} (a) and (b) show the training curve of CRP-SFT under different settings, demonstrating that the models pre-finetuned on DRP-SFT data have a faster convergence. Even though losses converge during fine-tuning across different settings, the model fine-tuned on DRP-SFT data achieves better final performance on the benchmarks. This shows that it is helpful to learn different types of visual operations separately before interleaving them during training. As shown in \Cref{tab:m4_vitevqa_app}, results on the validation set of M4-ViteVQA are also reported, demonstrating that Video-R4-7B establishes a new state-of-the-art in text-rich video understanding and reasoning.

\begin{table*}[!ht]
\centering
\caption{Performance comparison on the M4-ViteVQA validation set and testset.}
\label{tab:m4_vitevqa_app}
\resizebox{\textwidth}{!}{
\begin{tabular}{lcccccccccccc}
\toprule
\multirow{3}{*}{\textbf{Models}} &
\multicolumn{4}{c}{\textbf{Task 1 - Split 1}} &
\multicolumn{4}{c}{\textbf{Task 1 - Split 2}} &
\multicolumn{4}{c}{\textbf{Task 2}} \\
\cmidrule(lr){2-5}\cmidrule(lr){6-9}\cmidrule(lr){10-13}
& \multicolumn{2}{c}{Val} & \multicolumn{2}{c}{Test} &
  \multicolumn{2}{c}{Val} & \multicolumn{2}{c}{Test} &
  \multicolumn{2}{c}{Val} & \multicolumn{2}{c}{Test} \\
\cmidrule(lr){2-3}\cmidrule(lr){4-5}
\cmidrule(lr){6-7}\cmidrule(lr){8-9}
\cmidrule(lr){10-11}\cmidrule(lr){12-13}
& Acc.(\%) & ANLS(\%) & Acc.(\%) & ANLS(\%) &
  Acc.(\%) & ANLS(\%) & Acc.(\%) & ANLS(\%) &
  Acc.(\%) & ANLS(\%) & Acc.(\%) & ANLS(\%) \\
\midrule
JustAsk~\cite{yang2021justask}           &
10.81 & 15.40 & 10.05 & 14.10 &
7.16 & 10.00 & 5.47 & 8.60 &
4.86 & 6.70 & 3.60 & 6.70 \\

All-in-one-B~\cite{wang2023allinoneb}   &
11.47 & 15.30 & 10.87 & 14.80 &
6.85 & 9.20 & 5.66 & 7.80 &
4.20 & 5.00 & 3.28 & 4.60 \\

Video-LLaVA-7B~\cite{lin2024videollava} &
15.82 & 17.77 & 15.43 & 17.15 &
13.14 & 14.29 & 11.19 & 12.02 &
10.89 & 13.23 & 9.38 & 11.80 \\

T5-ViteVQA~\cite{zhao2022vitevqa}        &
23.17 & 30.10 & 22.17 & 29.10 &
17.59 & 23.10 & 16.68 & 23.80 &
12.30 & 16.10 & 9.29 & 13.60 \\

VideoLLaMA2-7B~\cite{cheng2024videollama2} &
20.04 & 21.73 & 20.76 & 23.55 &
18.30 & 19.63 & 18.33 & 20.45 &
19.68 & 23.62 & 16.54 & 21.80 \\

Qwen2-VL-7B~\cite{wang2024qwen2}         &
36.77 & 46.56 & 35.22 & 45.84 &
28.55 & 39.34 & 27.25 & 38.45 &
22.95 & 32.65 & 21.23 & 28.79 \\

TEA-L~\cite{zhang2025trackandanswer}     &
37.49 & 46.38 & 34.78 & 43.71 &
28.27 & 36.32 & 28.43 & 38.13 &
22.83 & 30.21 & 18.83 & 28.90 \\

NVILA-8B~\cite{liu2025nvila}             &
37.89 & 47.67 & 37.73 & 47.23 &
30.25 & 40.58 & 30.10 & 41.52 &
23.79 & 32.89 & 22.89 & 30.34 \\

GAT-L~\cite{zhang2025gatherandtrace}     &
38.01 & 47.53 & 38.30 & 48.23 &
31.35 & 41.33 & 30.90 & 41.81 &
24.54 & 33.30 & 22.13 & 30.75 \\

Qwen2.5-VL~\cite{bai2025qwen25}          &
22.22 & 48.67 & 26.53 & 44.91 &
17.84 & 46.72 & 24.34 & 39.60 &
22.31 & 42.21 & 32.81 & 50.82 \\

Video-R1-7B~\cite{feng2025videor1}       &
38.10 & 50.80 & 37.10 & 48.25 &
38.40 & 49.62 & 33.67 & 44.94 &
47.77 & 58.52 & 43.16 & 53.37 \\

Pixel-Reasoner~\cite{su2025pixelreasoner}&
54.44 & 63.57 & 52.91 & 61.44 &
54.69 & 62.58 & 48.88 & 58.23 &
63.78 & 69.93 & 58.97 & 65.32 \\

\rowcolor{blue!8}\textbf{Video-R4-7B (ours)} &
\textbf{57.33} & \textbf{66.92} & \textbf{56.17} & \textbf{65.22} &
\textbf{57.65} & \textbf{65.15} & \textbf{52.69} & \textbf{61.89} &
\textbf{69.03} & \textbf{75.45} & \textbf{64.21} & \textbf{69.99} \\

\midrule
Human &
-- & -- & 85.27 & 89.30 &
-- & -- & 78.41 & 82.80 &
-- & -- & 82.26 & 85.10 \\
\bottomrule
\end{tabular}
}
\vspace{-1em}
\end{table*}

\section{Training Details}
\label{app:sec:training_details}

For DRP-SFT, we use 7k data from Video-R4-CoT-17k for fine-tuning. The learning rate of $1\times 10^{-6}$ is adopted. We fully fine-tune the model instead of using LoRA~\cite{hu2022lora}.
For the RL after DRP-SFT, we use accuracy and the curiosity reward. There are 15k samples from Video-R4-RL-30k used during the stage. Following~\cite{su2025pixelreasoner}, the curiosity reward's hyperparameters are set as follows: $\alpha=0.5$, $\beta=0.05$, and $H=0.3$.
GRPO~\cite{guo2025deepseek} is adopted as the policy optimization method. Eight responses are sampled for each sample. 
For the CRP-SFT, 10k samples from Video-R4-CoT-10k are used, and other hyperparameters are the same as those in DRP-SFT.
For the RL after CRP-SFT, we adopt accuracy, diversity, representativeness, and curiosity reward, with the coefficients $\lambda_\text{div}=\lambda_\text{rep}=\lambda_\text{cur}=1$~\cite{zhou2018deep}.
All the models are trained on one H100 80G GPU.

% \begin{figure}[!ht]
%     \centering
%     \includegraphics[width=\linewidth]{fig/avg_res_length.png}
%     \caption{Caption}
%     \label{fig:avg_res_length}
% \end{figure}

% \begin{figure*}[!ht]
%     \centering
%     \includegraphics[width=\linewidth]{fig/bar.png}
%     \caption{Our Video-R4-7B model achieves state-of-the-art performance on the text-rich video understanding dataset M4-ViteVQA, and is also compatible with the LMMs with the same size.}
%     \label{fig:bar}
% \end{figure*}

\section{More Visualization Results}
\begin{figure*}[!ht]
    \centering
    \includegraphics[width=\linewidth]{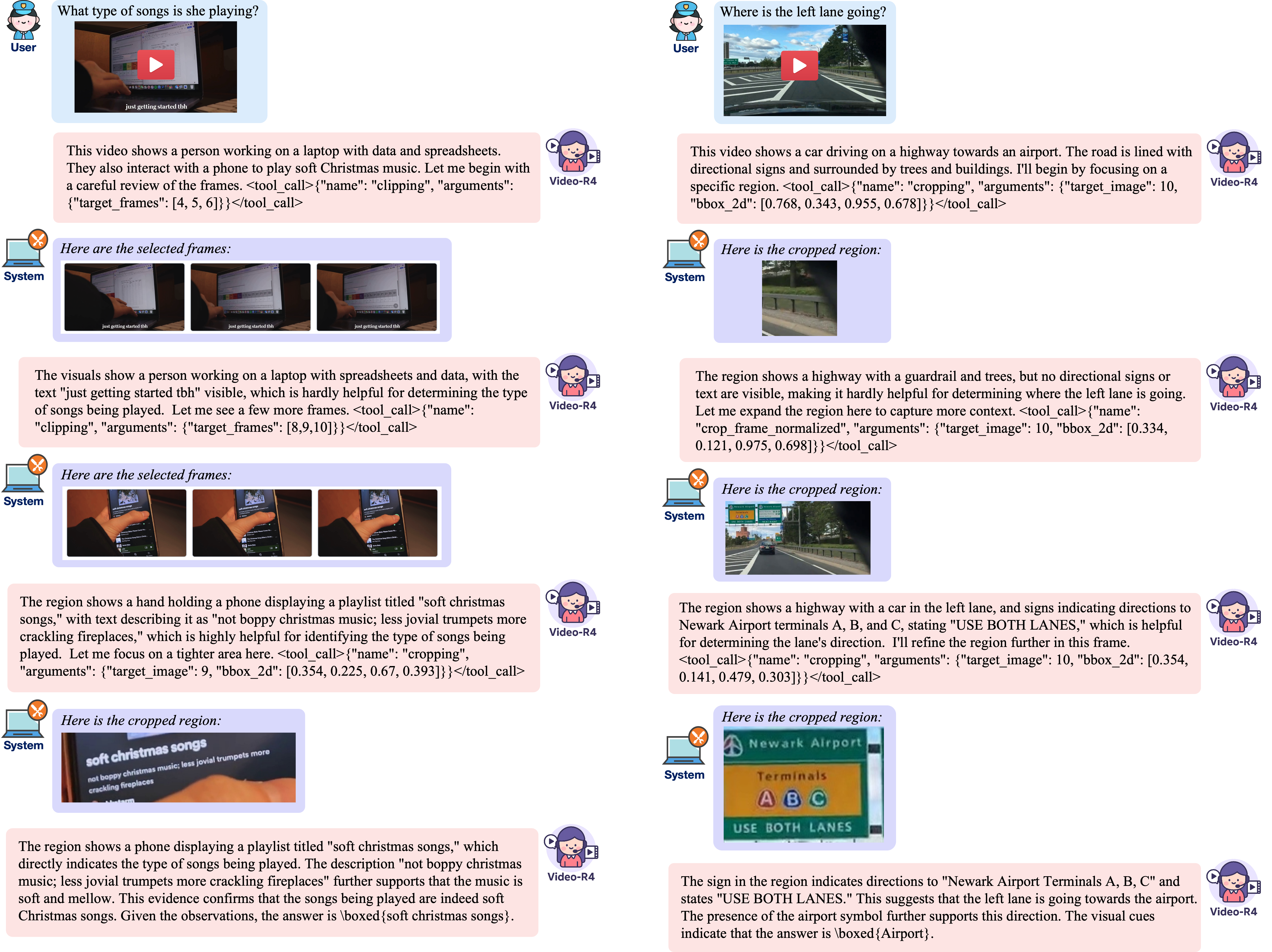}
    \caption{Trajectories visualization.}
    \label{fig:vis_1}
\end{figure*}

\begin{figure*}[!ht]
    \centering
    \includegraphics[width=\linewidth]{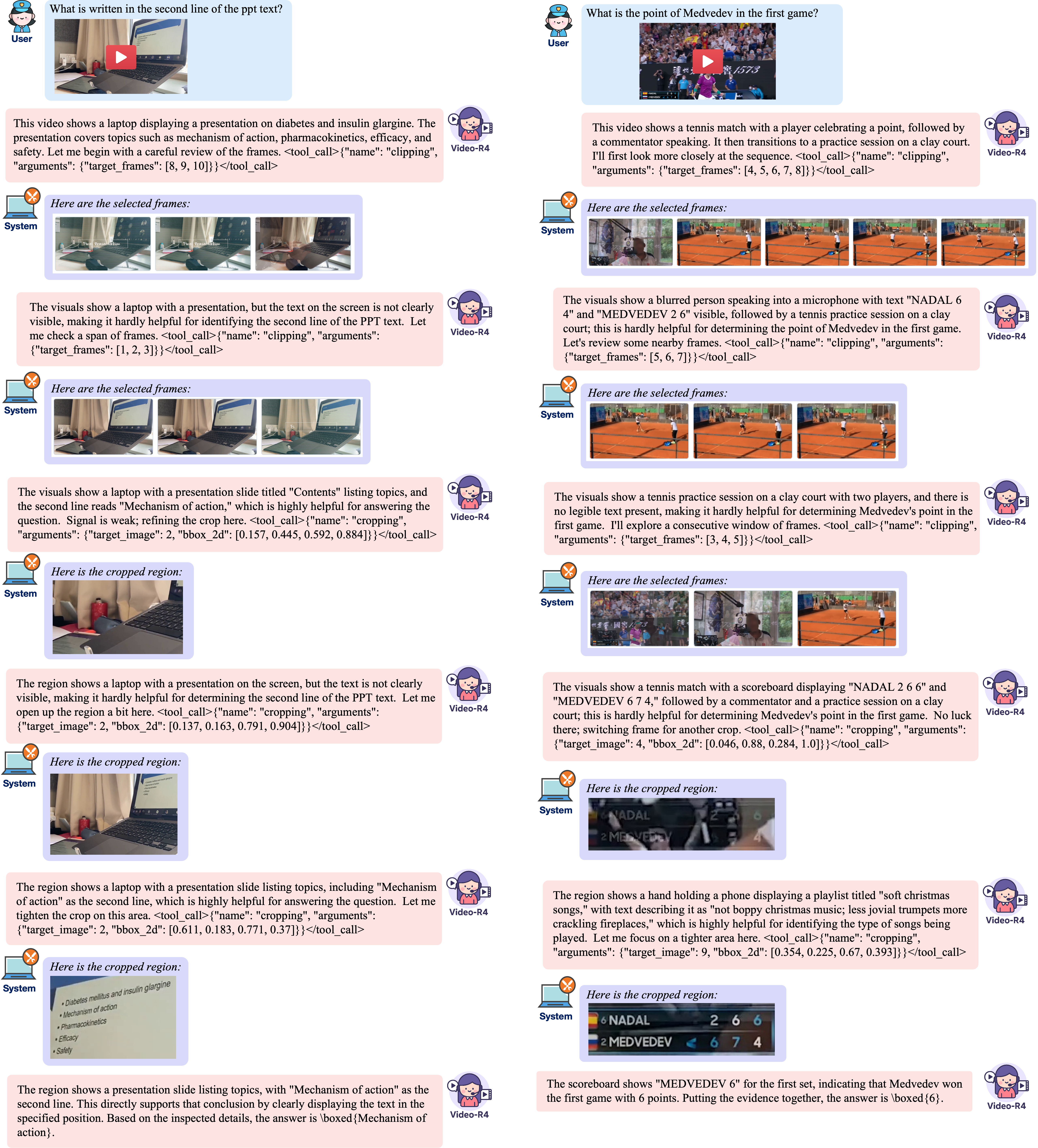}
    \caption{More visualization with longer trajectories.}
    \label{fig:vis_2}
\end{figure*}

As shown in \Cref{fig:vis_1,fig:vis_2}, we present additional visualizations of the trajectory samples.

\end{document}